\documentclass[twoside,11pt]{article}
\usepackage{blindtext}
\usepackage{jmlr2e}
\usepackage[english]{babel}
\usepackage[utf8]{inputenc}
\usepackage{subfig}
\usepackage{amsfonts}
\usepackage{amsmath}
\usepackage{comment}
\usepackage{moresize}
\usepackage{natbib}
\usepackage{float}

\usepackage{lastpage}
\jmlrheading{24}{2024}{1-\pageref{LastPage}}{1/24; Revised 3/24}{4/24}{21-0000}{Vahid Shamsaddini, Ali Ghofrani, Rahul Singh Inda, Jithendra Sai Veeramaneni, Étienne Voutaz and Wallyson Lemes de Oliveira}

\ShortHeadings{Graph Language Model (GLM): A new graph-based approach to detect social instabilities}{Shamsaddini, Ghofrani, Inda, Veeramaneni, Voutaz and Lemes de Oliveira}
\firstpageno{1}

\begin{document}

\title{Graph Language Model (GLM): A new graph-based approach to detect social instabilities}

\author{
\name Wallyson Lemes de Oliveira \email w.lemes@giotto.ai \\
       \addr Giotto.ai SA\\ Place de la Gare 4\\ 1004 Lausanne CH
\AND
\name Vahid Shamsaddini \email v.shamsaddini@giotto.ai \\
       \addr Giotto.ai SA\\ Place de la Gare 4\\ 1004 Lausanne CH
       \AND
       \name Ali Ghofrani \email a.ghofrani@giotto.ai \\
       \addr Giotto.ai SA\\ Place de la Gare 4\\ 1004 Lausanne CH
       \AND
       \name Rahul Singh Inda \email r.sinda@giotto.ai \\
       \addr Giotto.ai SA\\ Place de la Gare 4\\ 1004 Lausanne CH
       \AND
       \name Jithendra Sai Veeramaneni \email j.veeramaneni@giotto.ai \\
       \addr Giotto.ai SA\\ Place de la Gare 4\\ 1004 Lausanne CH
       \AND
       \name Étienne Voutaz \email etienne.voutaz@armasuisse.ch \\
       \addr Armasuisse\\ Feuerwerkerstrasse 39\\ 3602 Thun CH}

\editor{N/A}

\maketitle
\begin{abstract}%
This scientific report presents a novel methodology for the early prediction of important political events using News datasets. The methodology leverages natural language processing, graph theory, clique analysis, and semantic relationships to uncover hidden predictive signals within the data. Initially, we designed a preliminary version of the method and tested it on a few events. This analysis revealed limitations in the initial research phase. We then enhanced the model in two key ways: first, we added a filtration step to only consider politically relevant news before further processing; second, we adjusted the input features to make the alert system more sensitive to significant spikes in the data. After finalizing the improved methodology, we tested it on eleven events including US protests, the Ukraine war, and French protests. Results demonstrate the superiority of our approach compared to baseline methods. Through targeted refinements, our model can now provide earlier and more accurate predictions of major political events based on subtle patterns in news data.
\end{abstract}

\begin{keywords}
anomaly detection, graph theory, random graph theory, news data, war, protest, early warning signal
\end{keywords}

\section{Introduction}
The main problem that we will address in this project is the detection of early warning signals for important political events using News Datasets. Our objective is to investigate the feasibility of predicting events in advance. Previous studies -- like \citealp{chadefaux2014early}, \cite{mueller2018reading}, \cite{long2021news} and \cite{halkia2020conflict} -- have already explored the detection of wars using News Datasets. In this project, we aim to extend these works from two different perspectives:
\begin{itemize}
    \item \textbf{Predicting other social instabilities:} The first task is particularly challenging because other social instabilities, such as protests and riots, may not have as strong an impact on News Datasets compared to wars. Therefore, we need to overcome the difficulty of identifying subtle signals and patterns related to these events within the News Datasets using a combination of mathematics and machine learning.
    \item \textbf{Early prediction of wars:} The second task is also challenging due to the need to extract hidden patterns from the News Datasets that indicate the probability of events occurring. To achieve this, we will explore advanced techniques in mathematics to uncover these hidden patterns and improve the accuracy of early war prediction. 
\end{itemize}
To gain a deeper understanding of our primary concept, let's consider the Ukraine-Russia war as an example. If we aim to predict the outbreak of war just before it begins, the volume of news articles containing keywords such as "Ukraine," "Russia," and "war" would be significantly high, making it relatively straightforward to make predictions at that point. However, as we move further back in time from the start date of the war, the number of news articles directly mentioning Ukraine, Russia, and war decreases. Nonetheless, we may come across other news articles, such as one reporting that Biden contacted Putin regarding Ukraine's instability. In order to detect early signs of war, we need to understand the relationship between key figures like Putin and Russia, as well as the association between Biden and Ukraine. By identifying news articles that connect the concept of Ukraine (represented by Biden) with the concept of Russia (represented by Putin) and the concept of war, we can uncover valuable insights. Therefore, as per the explanation, we leverage two crucial elements: 1) relationships between entities present in the same news articles and 2) the identification of related concepts. Graph theory is employed to capture relationships, while NLP methods are used to determine if two words belong to the same concept. With this in mind, our algorithm is founded on NLP methods and graph theory, which form the cornerstone of our research approach.

\section{Dataset Description and Preprocessing}

To demonstrate the generalizability of our proposed approach, this study utilized datasets covering various events across different times and locations. We focused on several event types: large and small protests in the United States and France and the Russia-Ukraine war. For the US protests, we included all data within a one-year period. Similarly, for the Russia-Ukraine data, we considered an equivalent duration, and for the France protests, we analyzed an 18-month interval. Data specific to these events was collected using the \textit{GDELT API version 2}.
\paragraph{}

To capture every event, we gathered data that was relevant to its location and important country or cities name. For instance, for the Russia-Ukraine war, we targeted data related to Ukraine and Russia.
Utilizing the GDELT API version 2, we were able to collect high-scale data for analysis. In the subsequent sections, we will delve into the details of the queries used to fetch the data for each event and explain the rationale behind their selection.

\subsection{Fetching the Data}
The GDELT API version 2 (Global Database of Events, Language, and Tone) is a data service that provides access to a vast collection of global news articles and events. It offers a comprehensive platform for analyzing and understanding global events, covering a wide range of topics such as politics, conflicts, protests, and more. With the GDELT API version 2, we can access and retrieve high-scale data, enabling us to conduct in-depth analyses, monitor trends, and gain insights into global events. We used it for our project. 

\subsection{Protests within the US}
To collect data for protests in the United States, we implemented a filtering mechanism to retrieve news headlines associated with a state or city in the United States. In order to achieve this, we employed a GDELT query that instructed the GDELT API to return news articles containing the names of select US states, including California, New York, Texas, Pennsylvania, Virginia, Florida, and Massachusetts. We ensured that the articles were in English and originated from news sources within the United States. Although it was not feasible to include the names of all states due to character limitations in the GDELT API, we included the names of the most significant states. This filtering process allowed us to gather a focused dataset specifically linked to the regions of interest within the United States.
\begin{figure}[H]
\centering
\includegraphics[width=1\textwidth]{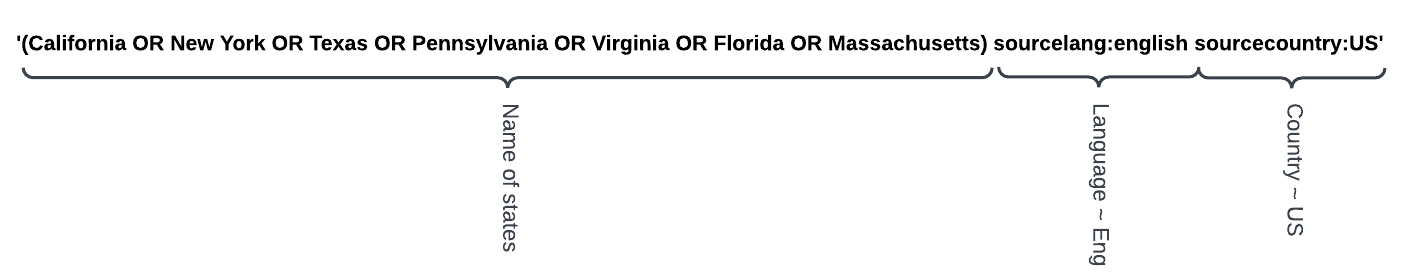}
\caption{\label{fig1}Example of GDELT query to fetch US protest related news.}
\end{figure}
After collecting the data using the aforementioned query, we further refined our dataset by applying string matching techniques to the news headlines. Specifically, we focused on cases where the headlines contained the names of at least one of the cities or states within the United States. Here are some examples of our fetched news:

\begin{table}[H]
\centering 
\begin{tabular}{|p{12cm}||p{3cm}|} 
\hline
\textbf{News Title} & \textbf{Published Date} \\
\hline

Sunnyvale residents join protest against airplane noise at San Carlos Airport – The Mercury News & 2017-06-30 \\
 \hline
Dallas Police Department Faces Struggles After Last Year Deadly Ambush & 2017-07-03  \\
 \hline
Trans Activists Protest Trump Transgender Military Ban in New York, San Francisco, and D.C.  & 2017-07-27  \\
 \hline
 Arizona teachers prepare for walk-in demonstrations over pay-Herald-Whig & 2018-04-12 \\ 
 \hline
 
\end{tabular}
\caption{\label{tab-hashtags}GDELT data US protest samples}
\end{table}

\subsection{Russia-Ukraine War}
To focus specifically on the Russia-Ukraine war, we utilized the GDELT API without encountering character limitations as we just need to consider "Russia" and "Ukraine" terms. By employing the query parameters "repeat8:Ukraine" and "repeat8:Russia," we ensured that the retrieved news headlines contained the terms "Ukraine" and "Russia" at least eight times. By leveraging this criterion, we obtained a set of news articles that were highly relevant and provided comprehensive coverage of the Russia and Ukraine problems:
\begin{figure}[H]
\centering
\includegraphics[width=0.42\textwidth]{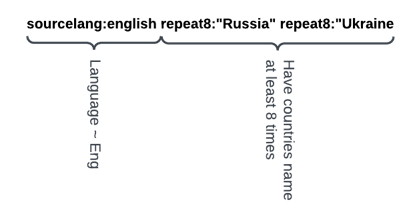}
\caption{\label{fig2}Example of GDELT query to fetch Ukraine war-related news.}
\end{figure}
In the case of the Russia-Ukraine war, we ensured that the collected news articles contained at least one of the "Ukraine", "Ukrainian", "Russia" and "Russian" keywords in their headlines. The table below shows some of the fetched news:
\begin{table}[H]
\centering 
\begin{tabular}{|p{12cm}||p{3cm}|} 
\hline
\textbf{News Title} & \textbf{Published Date} \\
\hline

Ukrainian president leaves Washington buoyed by Trump support & 2017-06-23 \\
 \hline
Experts Suspect Russia Is Using Ukraine As A Cyberwar Testing Ground & 2017-06-23  \\
 \hline
Ukraine reclaims full control from Russia of logistics hub, expects more gains & 2022-03-10  \\
 \hline
Biden praises Denmark for  standing up  for Ukraine in war with Russia & 2023-06-25  \\
 \hline
When Mom Believes Putin: A Russian Family Torn Apart Over Ukraine Invasion  & 2023-06-25  \\
 \hline
 
\end{tabular}
\caption{\label{tab-hashtags2}GDELT data Ukraine war samples}
\end{table}

\subsection{French Protests and Riots}

 To collect data for French Riots and Protests, we implemented a filtering mechanism to retrieve news articles associated with the city names in France. To achieve this, we employed a GDELT query instructing the GDELT API to return articles containing the names of key French cities, including Paris, Marseille, Lyon, Toulouse, Nice, Nantes, Montpellier, Strasbourg, Bordeaux, and Lille, we also included the terms 'France' and 'French'. We ensured the articles were in English, but for this event, we did not limit ourselves to the source country of France because of the significance of the international community's connection to these events, we chose not to specify the source country, allowing us to gather French-related news from all over the world.
\begin{figure}[H]
\centering
\includegraphics[width=1\textwidth]{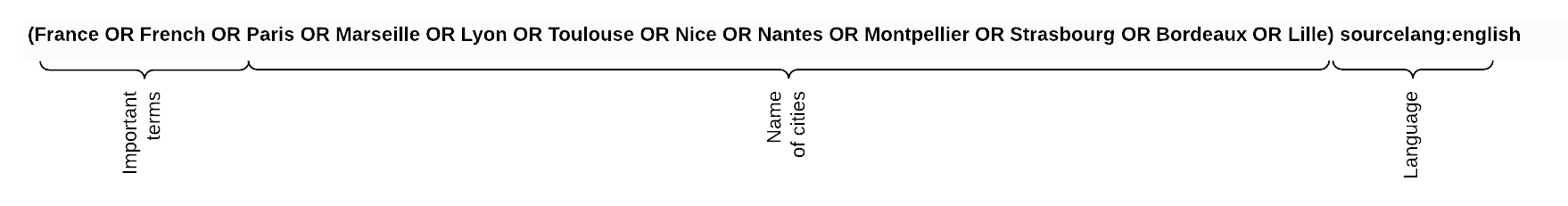}
\caption{\label{fig4}Example of GDELT query to fetch French Riots related news.}
\end{figure}
Here are some examples of our fetched news:

\begin{table}[H]
\centering 
\begin{tabular}{|p{12cm}||p{3cm}|} 
\hline
\textbf{News Title} & \textbf{Published Date} \\
\hline

France vs Morocco: Hakimi ready to face Mbappe in World Cup semis & 2022-12-11 \\
 \hline
World's most powerful tourism cities: Paris tops, Is any Indian city on the list? Check here & 2023-01-20  \\
 \hline
French Yellow Vests Celebrate One Year Anniversary As General Strike Looms & 2019-11-20  \\
 \hline
France to grind to a halt : Furious transport workers to strike over Macron pension reforms & 2019-11-25  \\
 \hline
COVID - 19 : First French citizen dies as German man remains in critical condition  & 2020-02-26  \\
 \hline
 
\end{tabular}
\caption{\label{tab-hashtags3}GDELT data French Riots samples}
\end{table}

\section{Methodology:}

In this section, we describe our novel method, the \textbf{Graph Language Model (GLM)}, and its subsequent improvements.

\subsection{Graph Language Model (GLM v1)}
\begin{figure}[H]
\centering
\includegraphics[width=0.95\textwidth]{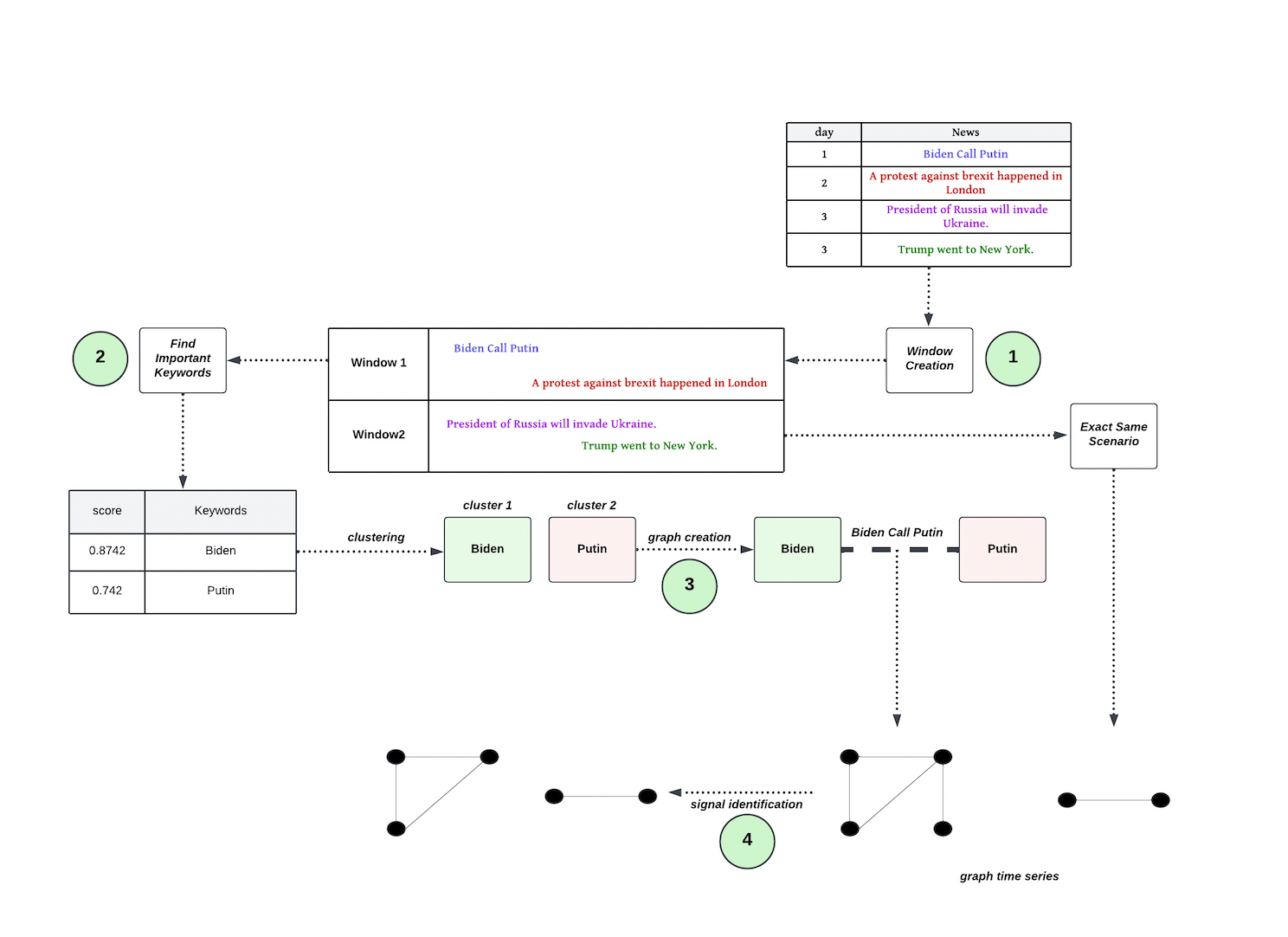}
\caption{\label{fig_pipeline}The Overall Pipeline of the method, Part 1}
\end{figure}

In our new method, \textbf{Graph Language Model (GLM v1)}, initially, we create a time series of news by collecting articles over time (1st arrow of Figure \ref{fig_pipeline}). Then, we find important keywords in each collection by ranking them via EmbedRank (2nd arrow of Figure \ref{fig_pipeline}). Once it has identified different clusters of keywords, we create a graph with the different clusters as its nodes and with edges showing the coexistence of two clusters within a single article (3rd
arrow of Figure \ref{fig_pipeline}). Subsequently, we will consider cliques as signals and explore them (4th arrow of Figure \ref{fig_pipeline}). 

\begin{figure}[H]
\centering
\includegraphics[width=0.75\textwidth]{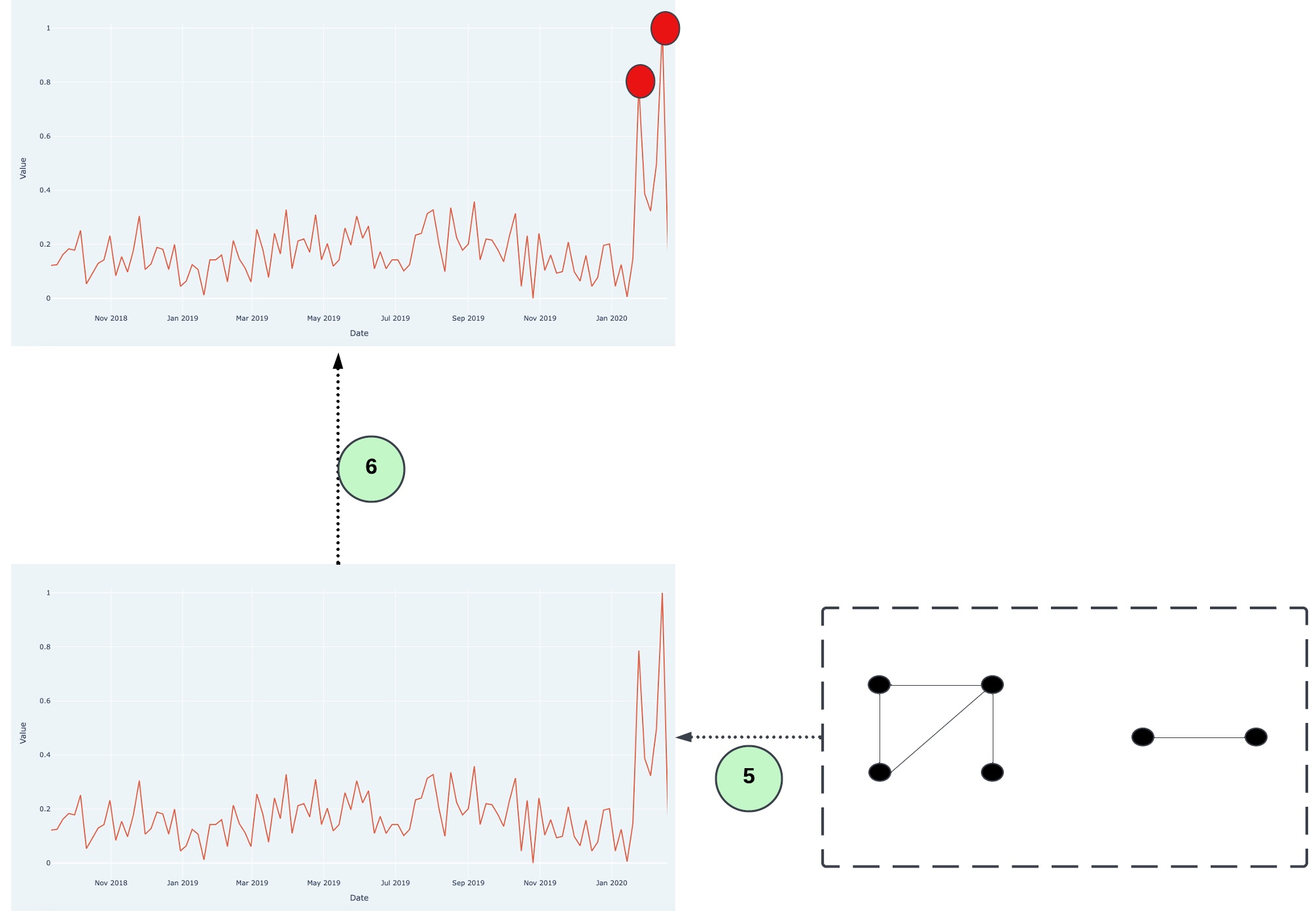}
\caption{\label{fig_pipeline_part2}The Overall Pipeline of the method, Part 2}
\end{figure}

Afterward, we leverage the time series of the graph to extract informative features using mathematical tools,
enabling us to monitor the evolution of the graph and signals. These informative features are transformed into
time series data (5th arrow of Figure \ref{fig_pipeline_part2}). Finally, we utilize this time series data (and some lags of them) as input for the Alert system,
which is designed to detect anomalies (6th arrow of Figure \ref{fig_pipeline_part2}).  In the upcoming sections, we will delve into the specifics of each part in greater detail.

\subsubsection{Window Creation}
The initial step of our method involves partitioning our news dataset, taking into consideration the temporal nature of the data. We divide the entire dataset into smaller segments, referred to as \textbf{windows}, which have equal lengths. The length of the windows is a hyperparameter of the model. Moreover, we introduce an overlap between consecutive pairs of windows, where they share a certain number of common days, this was done to more effectively track and understand the transitions that occur between these periods. This overlap parameter is also a hyperparameter in our methodology. For instance, in Figure \ref{fig_pipeline} the window lengths are two days and the intersection is one day.

\subsubsection{Finding Keywords}
The subsequent step in our methodology involves identifying the important keywords within each news article. This step comprises two distinct parts: firstly, the identification of all potential keywords, and secondly, the selection of the most relevant keywords. 
\begin{figure}[H]
\centering
\includegraphics[width=0.65\textwidth]{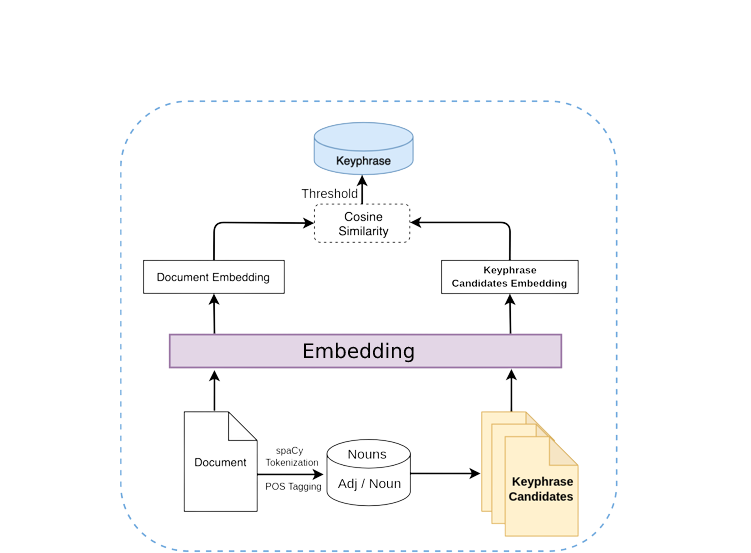}
\caption{\label{fig_keyword_extract_ali1}The Overall Pipeline of the keyword extraction.}
\end{figure}

Here are the steps involving the Keyword extraction pipeline.

\begin{enumerate}
\item Tokenize the news headlines using the spaCy tokenizer to break them down into individual words.

\item Extract keyword candidates from the headlines by identifying sequences of adjectives that end with a noun. These candidates are identified using the AdjNoun/Noun POS tags.

\item Extract embedding features for both the keyword candidates and the news headlines using the state-of-the-art Sentence Transformer-based MPnet model. This allows us to capture the semantic meaning and context of the keywords and headlines, and to identify patterns and relationships within the data. (see \cite{song2020mpnet, reimers2019sentence})

\item Compute the similarity between each keyword candidate and the news headline. This allows us to identify the most important keywords that are most closely related to the headline.

\item Apply a threshold to the keyword candidates similarity scores to filter out less relevant keywords and ensure that only the most important ones are included in our analysis.
\end{enumerate}

\subsubsection{Topics Formation (Clustering)}
Once we have identified all the important keywords from different news articles, the next step is to cluster these keywords within each window. For clustering, we employ a two-step approach. Firstly, we use \textit{Sentence Transformer} for embedding the important keywords, creating numerical representations of the keyword semantics. Then, we apply the \textit{Fastcluster algorithm} (see \cite{mullner2013fastcluster}) to cluster these embedded keywords.

\paragraph{}
By clustering the keywords, we aim to group together related terms. For example, "Putin" and "Vladimir Putin" would be considered part of the same cluster. This allows us to capture semantic similarities among the keywords. 
\begin{description}
\item[Topic:] In the context of our paper, a "topic" is defined as a cluster of related terms, each conveying a similar meaning or having a semantic relation.
\end{description}

\subsubsection{Graph Creation}
To capture the relationships between different clusters of keywords, we employed a graph-based approach. In this methodology, each cluster of keywords was represented as a node in the graph. To capture the co-occurrence of clusters within news headlines, we established edges between nodes.

By creating these edges, we were able to identify clusters that frequently appeared together in news articles, indicating a potential relationship or connection between the two clusters. This graph-based representation allowed us to analyze the interconnections between keyword clusters and uncover meaningful patterns and associations within the dataset.

\begin{figure}[H]
\centering
\includegraphics[width=0.6\textwidth]{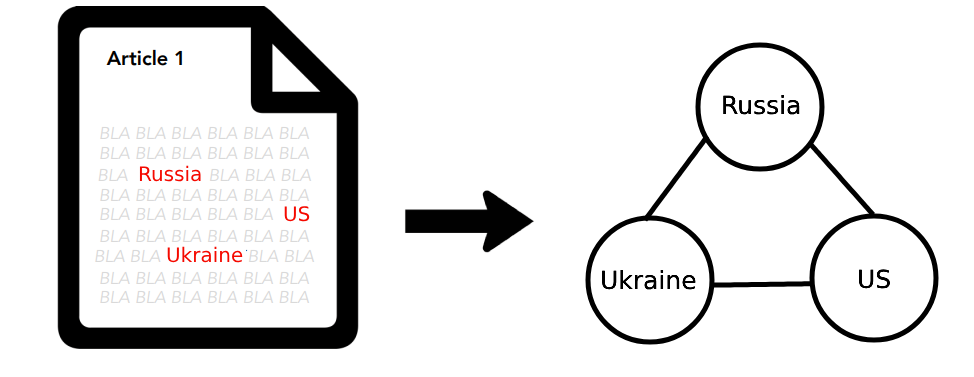}
\caption{\label{graph_example}The Overall Pipeline of the Graph Creation}
\end{figure}

\subsubsection{Signal Identification}
 In our methodology, we consider cliques as signals within our graphs. A clique refers to a group of nodes where every pair of nodes within the group is interconnected. The emergence of significant events is expected to manifest as discernible patterns in the graphs we have designed, generating interconnected clusters of topics. We hypothesize that these patterns materialize in the form of cliques. 
\paragraph{}
To illustrate this, let's consider the example of the Ukraine-Russia war. When we apply clustering to a window near the start date of the war, we typically obtain a cluster representing Ukraine and related terms, another cluster representing Russia, and potentially a cluster representing the United States or other countries with significant involvement in the conflict.  In the graph representation, the occurrence of news articles mentioning all three countries would result in a clique that contains these three nodes (see the clique with three nodes in fig \ref{graph_example}). Our assumption is that those cliques are likely to indicate upcoming important events, with each node representing a different topic of the clique.
\paragraph{}
By identifying and analyzing cliques within our graphs, we aim to uncover meaningful patterns and relationships that can serve as signals for significant events. 

\subsubsection{Feature Extraction}
In our methodology, we extract informative features from the time series of the graph and the identified signals. These features are derived using mathematical tools to monitor the evolution of the graph and signals.

\paragraph{}
In this project, we focus on two important features that capture the information of signals:

\begin{itemize}
    \item The first feature we consider is the \textbf{number of cliques}, which measures the count of cliques detected within the graph compared to a completely random graph. This feature is based on the assumption that in the absence of important events, the graph structure would resemble a random graph. By estimating the expected number of cliques in an Erdos-Renyi random graph, we can evaluate the deviation from this expectation and identify potential significant events.

    According to \cite{sakurai2022counting}, the expected number of cliques in an Erdos-Renyi random graph is given by:
        $$\mathbb{E}\left[X_{n, p}\right]=n^{\frac{1}{-2 \log p}(\log n-2 \log \log n+O(1))}$$
    Here, $n$ represents the number of vertices (nodes) in the graph, and $p$ represents the edge probability parameter. To estimate $n$ and $p$ for our graph, we assume that our graph follows a random structure similar to an Erdos-Renyi graph. The value of $n$ can be obtained directly as the number of nodes, while $p$ can be estimated as $\frac{m}{\left(\begin{array}{l}
n \\
2
\end{array}\right)}$, where $m$ is the number of edges in the graph.

At each window, we calculate the difference between the expected number of cliques and the actual number of cliques detected in the graph. This provides us with an indication of potential deviations from a random structure and allows us to identify significant events or deviations from the expected pattern.

\item Another feature we consider is the \textbf{heaviness of cliques}, which defines the mean of degrees of clique nodes (see fig \ref{fig_heaviness}).
 It is clear that the result will be at least $N_{nodes}-1$ where $N_{nodes}$ is the number of nodes within the clique. To compute this feature for a window, firstly, we calculate heaviness for each clique within the window, so we have a collection of clique heaviness, and subsequently, to have a single feature value for entire window, we consider the average of (or possibly the maximum or minimum) heaviness values of all cliques in the graph. 

\begin{figure}[H]
\centering
\includegraphics[width=0.5\textwidth]{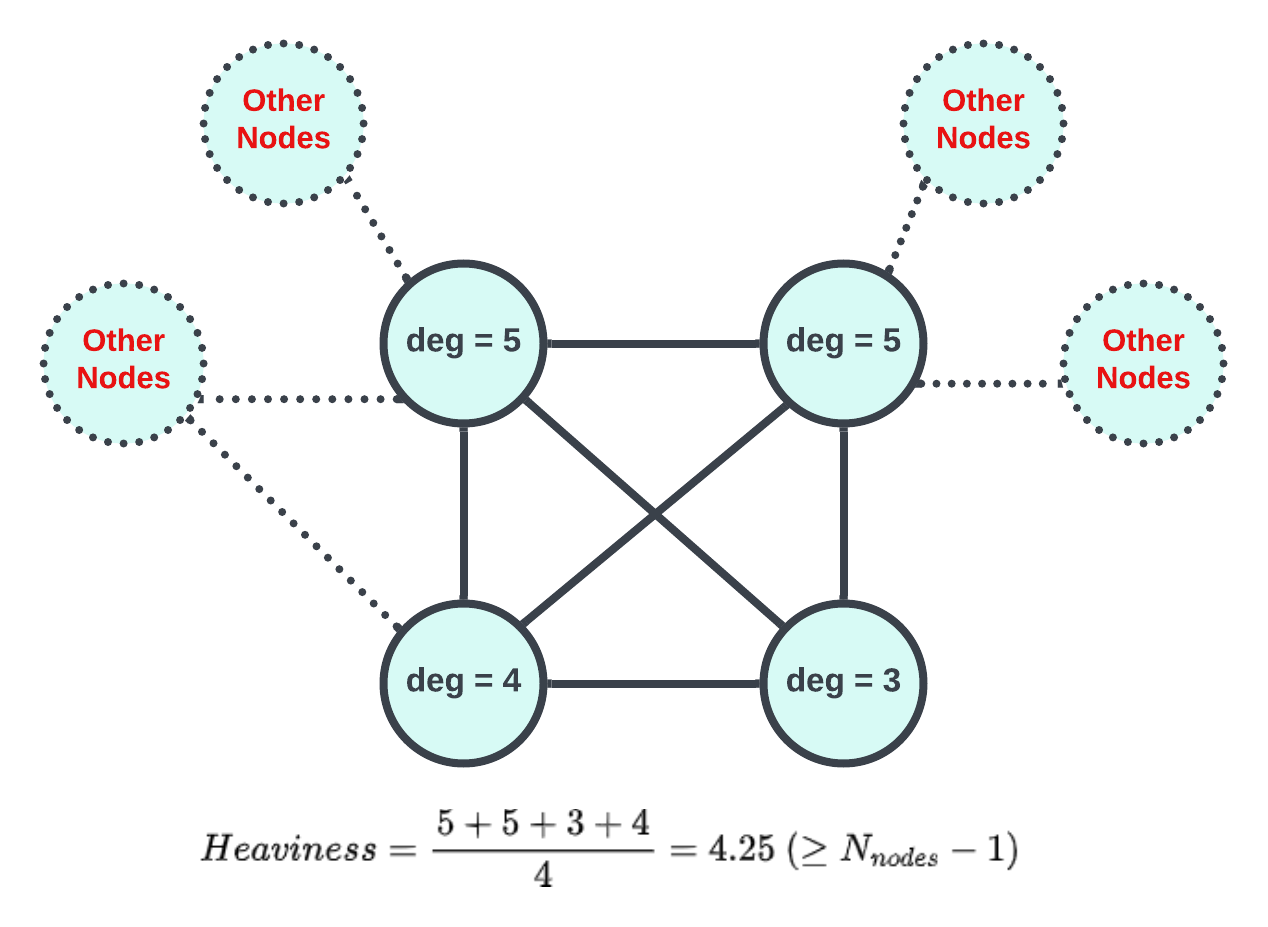}
\caption{\label{fig_heaviness}Heaviness of cliques}
\end{figure}

In our method, the heaviness of a clique serves as a reflection of the strength or importance of the connections within the graph. A high heaviness value indicates that the clique is a central hub with numerous connections to other nodes. By analyzing the average heaviness across all cliques, we can gain insights into the overall intensity or density of the interactions captured by the cliques. This feature provides a measure of the collective impact or significance of the identified events within the graph.

By tracking the heaviness of cliques over time, we can observe variations in the strength of interactions and identify periods of increased or decreased connectivity. This allows us to detect and analyze changes in the graph dynamics.  
\end{itemize}
\subsubsection{Alert System}
    In our methodology, we utilize the defined time series and their lagged histories as features for each window. We incorporate a short history of these feature sets for the input of the isolation forest algorithm. The isolation forest is an algorithm for data anomaly detection initially developed by \cite{liu2008isolation}. Isolation Forest detects anomalies using binary trees. The algorithm has a linear time complexity and a low memory requirement, which works well with high-volume data. In essence, the algorithm performs a fast approximate density estimation, and considers points with a low density estimate as anomalies. To help intuitively understand its workings, consider these two fundamental concepts:
\begin{itemize}
    \item \textbf{Shorter Path Length}: Anomalies are typically isolated using fewer splits, indicating that unusual data points can be differentiated from the rest more swiftly.
     \item \textbf{Anomaly Score}: The average path length, which corresponds to the number of splits needed to isolate a point, is used to gauge the degree of anomaly. Lower scores imply more anomalous data points.
\end{itemize} 
\begin{figure}[H]
\centering
\includegraphics[width=0.80\textwidth]{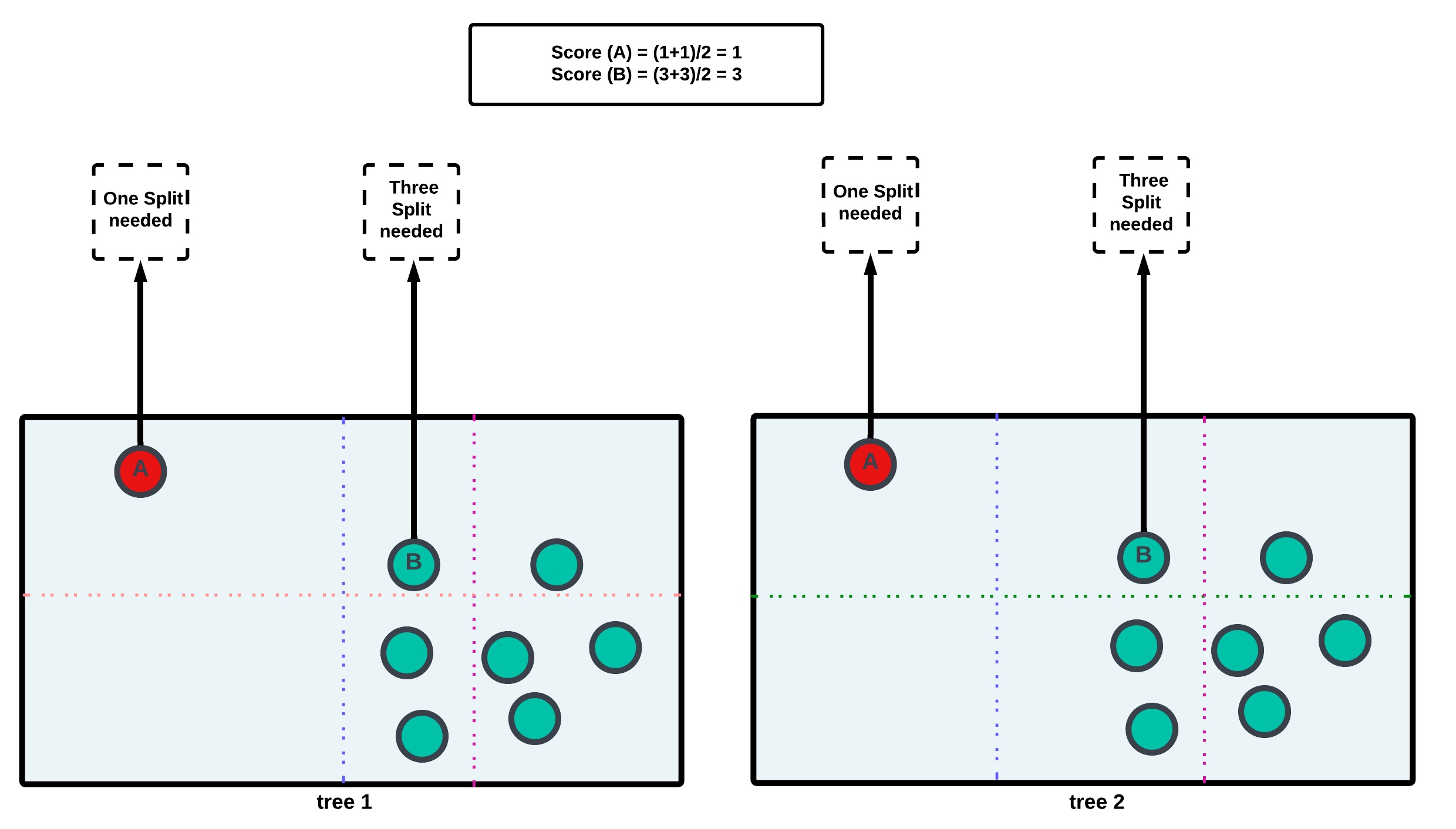}
\caption{\label{fig_iforest}The logic of anomaly score of iForest}
\end{figure}

In other words, the unique aspect of the Isolation Forest algorithm is that it isolates anomalies instead of the most common data points, thus enabling a fast and effective method for anomaly detection.

Here are the steps involving the Random Forest algorithm.

\begin{enumerate}

\item \textbf {Randomly select a subset of the data:} The algorithm randomly selects a subset of the data to create an isolation tree. The subset is chosen by randomly selecting a feature and then randomly selecting a split value within the range of the feature's values.

\item \textbf {Continue splitting until the data points are isolated:} The algorithm continues to split the data subset by randomly selecting features and splitting values until each data point is isolated in its own leaf node.

\item \textbf {Create multiple trees:} The algorithm repeats steps 1 and 2 to create multiple isolation trees.

\item \textbf {Calculate the anomaly score:} For each data point, the algorithm calculates an anomaly score based on the average path length of that data point in all the isolation trees. The anomaly score represents how isolated or how different the data point is from the rest of the data.

\item \textbf {Determine the outliers:} The algorithm determines which data points are outliers by comparing their anomaly scores to a threshold value. Data points with scores above the threshold are considered outliers.
\end{enumerate}

To facilitate the interpretation of the anomaly scores, we scale the scores between 0 and 1 using a min-max scaler. This transformation allows us to map the anomaly scores to a consistent range. We then set a threshold between 0 and 1, which serves as a criterion for classifying samples as anomalies. If the anomaly score of a sample exceeds this threshold, we consider it as an anomaly.

\subsubsection{Disadvantages of GLM v1:}
In the second phase, after a thorough analysis of data and outputs across various datasets, we identified two major limitations of the initial GLM approach:

\begin{enumerate}
    \item \label{first_dis} \textbf{Indiscriminate Signals:} The methodology did not distinguish between events relevant to the prediction task versus irrelevant incidents. For example, cliques with high edge weights may represent events unrelated to political protests or wars, such as news about an archeological discovery. For instance, consider the clique that Figure \ref{indiscriminate-signals-to-tex} shows:
    \begin{figure}[H]
    \centering
    \includegraphics[width=0.3\textwidth]{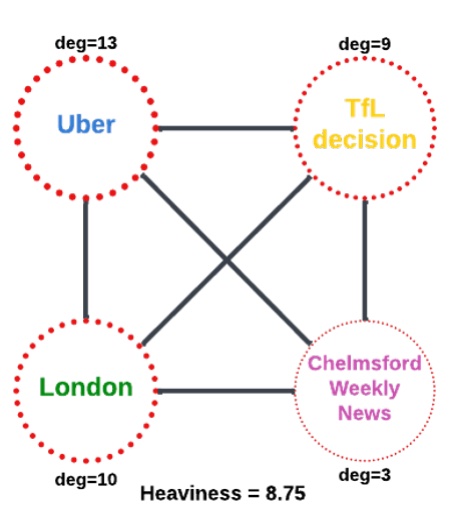}
    \caption{\label{indiscriminate-signals-to-tex} An irrelevant signal}
    \end{figure}
    This can lead to high false positive rates as irrelevant events trigger alerts. It can also increase false negatives by 'drowning out' relevant signals. For instance, if a protest coincides with the World Cup final, the model may only detect the sporting event due to the overwhelming volume of associated news.

    \item \label{second_dis} \textbf{Poor Feature Extraction:} The initial feature set (time series and their lags) used in the time series alert system was inadequate to detect early signals before event spikes emerged. Key predictive indicators were likely overlooked, limiting the system's ability to issue timely warnings.
\end{enumerate}
In the following sections, we detail the solutions developed to address these limitations through more selective graph construction and an expanded, optimized feature set for the alert system. Resolving these disadvantages significantly improved the methodology's accuracy, precision, and lead time compared to the initial approach.

\subsection{GLM v2}
\begin{figure}[H]
\centering
\includegraphics[width=0.95\textwidth]{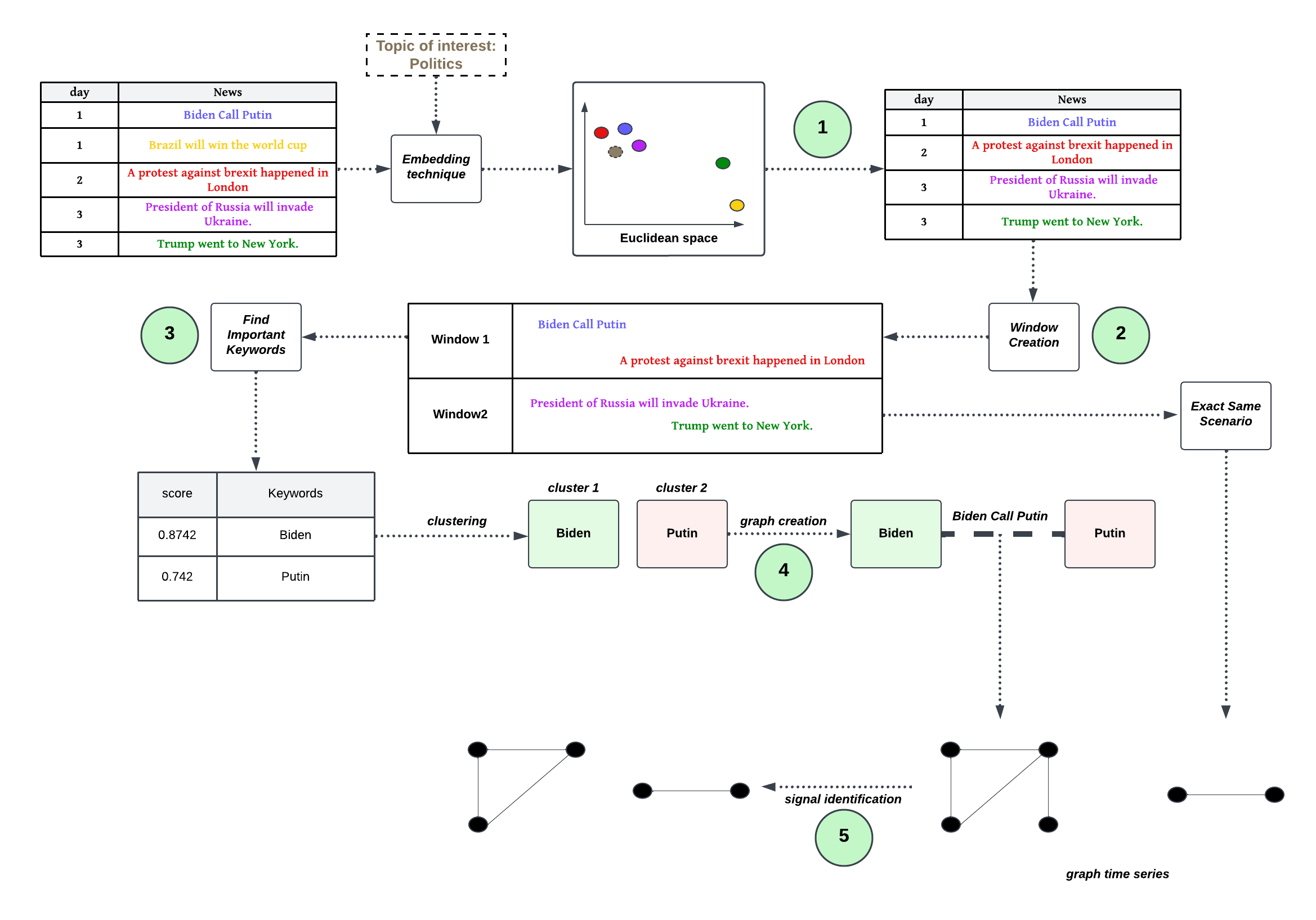}
\caption{\label{fig_glm2}Enchanced model (GLM v2)}
\end{figure}

Building upon the foundation laid by \textbf{GLM v1}, our enhanced model, \textbf{GLM v2}, introduces a preliminary step of filtering based on semantic similarity to exclude news articles irrelevant to our topic of interest, addressing the issue of indiscriminate signals (see disadvantage \ref{first_dis}) and refining the input dataset (1st arrow of Figure \ref{fig_glm2}).

The refined collection process streamlines the subsequent steps of constructing the time series of news, extracting keywords, forming clusters, and graph creation (2nd to 4th arrows of Figure \ref{fig_glm2}). GLM v2 maintains the use of cliques as signals (5th arrow of Figure \ref{fig_glm2}) and follows the same subsequent process as GLM to analyze the graph's time series and generate time series data (like 5th arrow of Figure \ref{fig_pipeline_part2}). To address the second weakness (see disadvantage \ref{second_dis}), the Alert system in GLM v2 also benefits from the refined features that provide a denoised representation of the time series, which allows our Alert system to better capture abrupt transitions (like 6th arrow of Figure \ref{fig_pipeline_part2}).  A deeper dive into the specific advancements and modifications introduced in GLM v2 will be presented in the forthcoming sections.

\subsubsection{Filtration}
To filter out news articles that are not relevant to our topic of interest, we utilize the \textit{bge-large-en-v1.5} text embedding from FlagEmbedding (see \cite{bge_embedding}). This embedding has achieved state-of-the-art (SOTA) results through October 2023 in the Massive Text Embedding Benchmark (MTEB) challenges (see \cite{muennighoff2023mteb}).


We obtain the embeddings for every news headline as well as for our topic of interest in this case, 'Politics'. We then filter out news articles based on the cosine similarity between their embeddings and the topic embedding, retaining only those articles whose similarity exceeds a fixed threshold of 0.4.

To determine a suitable threshold, we utilize the News category dataset from \cite{misra2022news}. This dataset includes a category column indicating whether a news article is related to politics or not. From this dataset, we randomly sampled 100 politics-related news articles and 100 comedy news articles. We then computed the cosine similarity between the embeddings of these 200 news articles and the embedding of 'Politics', resulting in the following histogram:
\begin{figure}[H]
\centering
\includegraphics[width=0.35\textwidth]{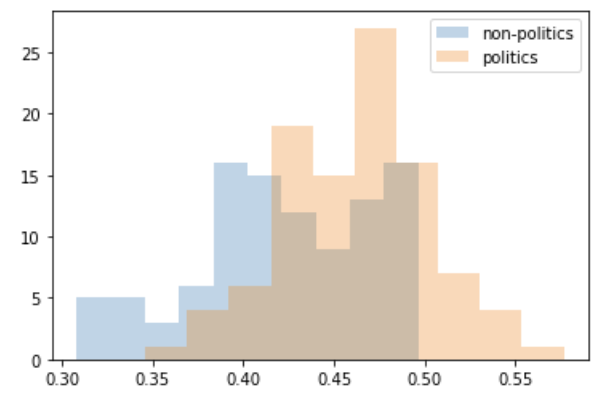}
\caption{\label{hist_cosine_sim} histogram of cosine similarities}
\end{figure}

We set the threshold to 0.4 as our primary objective was to identify all relevant events without missing any. Furthermore, to ascertain the efficacy of using this embedding coupled with cosine similarity for filtering, we conducted a t-test to examine if a significant difference exists between the means of cosine similarity values for politics-related and non-politics news articles. The p-value obtained was 1.00e-19, indicating a statistically significant difference and suggesting that this method is viable. We conducted a similar test using another embedding—the Sentence Transformer-based MPnet model. However, in this case, the p-value was 0.59, demonstrating that this alternative embedding is not suitable for our application.

\subsubsection{Alert System}

For this section, we just change the \textbf{Time Series Feature Extraction} part to solve the disadvantage \ref{second_dis} of \textbf{GLM v2}. Feature extraction is a crucial step for time series anomaly detection as it allows the transformation of raw time series data into a more informative representation that better captures patterns and anomalies.

\paragraph{}

In the original \textbf{Graph Language Model (GLM)} implementation, feature extraction was limited, relying primarily on the time series of clique count and heaviness along with their lagged values as input for the alert system. While providing useful signals, this approach may overlook more subtle anomalies not directly captured by those indicators alone. 

\paragraph{}

Contrastingly, \textbf{GLM v2} introduces a more sophisticated method for feature extraction, significantly enhancing the model's capacity to discern subtle and complex patterns within the time series data. Rather than solely depending on the raw count and weight of cliques, \textbf{GLM v2} expands the feature dimensions by incorporating a range of statistical functions calculated from a short history within the time series.

\paragraph{}

To elucidate, we integrate a hyperparameter $L_{f}$, representing the number of lags considered for feature extraction. Once $L_{f}$ is chosen, for each step $t$, the time series values at steps $t, t-1, ..., t-L_{f}$ are taken into account, and statistics are extracted from these values as a feature for step $t$. In all our experiments, $L_{f}$ (called Lag of time series feature extraction in Result section) is set to 15, and we utilize \textit{max}, \textit{median}, and 90th percentile as features. Additionally, we incorporate the value at step $t$ as a feature. Therefore, the feature space for each window contains eight values (Four features for each time series)
: \textit{max}, \textit{median}, and 90th percentile of a short history and time series itself. Once we generate the eight-dimensional features for each window, forming new representations, we also incorporate a short history of these representations for the input of the alert system. This inclusion involves a new lag, which we refer to as the "Lag of Alert System" in result section.
\paragraph{}

This multi-scale feature representation equips the anomaly detection model with a more comprehensive perspective on the temporal patterns embedded in the data, enabling a more effective distinction between normal fluctuations and anomalous trends, even when these anomalies occur over extended durations. The features extracted in \textbf{GLM v2}, therefore, provide a richer and more discriminating set of inputs.

\section{Baseline approach}

To evaluate the performance of our method and compare it with a common approach, we conducted a benchmarking analysis using information from \cite{chadefaux2014early}. The baseline approach we considered involved using the time series data of the number of news articles containing at least one event-related term. By event-related we mean:

\begin{itemize}
\item \textbf{For War}: conflict, war, battle, crisis, clash, fight, attack, combat, struggle, fighting, confrontation.
\item \textbf{For Protests}: protest, protester.
\end{itemize}

For the baseline, we simply consider the mentioned time series and 7 of its lags, and then apply the same alert system as GLM. By evaluating the results of both \textbf{baseline} and \textbf{GLM v2}, we can assess the effectiveness and potential advantages of our method in detecting and predicting important events compared to the traditional approach based on the number of event-related news articles.

\section{Results and Discussion}


In this section, we present our results benchmarked against the baseline approach. For all results, we use the same hyperparameters as reported in the following table (unless otherwise specified):

\begin{table}[H]
\centering 
\begin{tabular}{|p{5.8cm}||p{3cm}|p{3cm}|} 
\hline
\textbf{Hyperparameters} & \textbf{Value} \\
\hline

Window Size & 7\\
 \hline
Intersection Size & 5\\
\hline
Anomaly Threshold & 0.80\\
\hline
Lag of time series feature extraction & 15\\
\hline
Lag of Alert system & 15\\
\hline
\end{tabular}
\caption{\label{tab-hyperpar_ukraine}Hyperparameters}
\end{table}
In the following sections, we detail the various results. In this section, "GLM" refers to "GLM v2", unless otherwise specified. 

\paragraph{Performance Metrics Definitions:}
\begin{itemize}
\item \textbf{True Positive (TP):}
A trigger is classified as a True Positive if it is event-related and occurs prior to the event in case of single-date events, or before the end date for multiple-date events.

\item \textbf{False Positive (FP):}
A trigger is designated as a False Positive when it is issued by the algorithm without a subsequent occurrence of a related event.

\item \textbf{False Negative (FN):}
A trigger is classified as a False Negative if it is event-related but is issued after the occurrence of a single-date event or after the end date of a multiple-date event.

\item \textbf{Misses:}
An event is said to be missed if no \textbf{TP} trigger is issued by the algorithm.

\item \textbf{Detections:}
An event is said to be detected if a \textbf{TP} trigger is issued by the algorithm.


\end{itemize}
   
We consider triggers as event-related if the heaviest cliques were relevant to the event, or if a significant number of the cliques expressed news relevant to that event. For the baseline, a trigger is considered event-related only if a major portion of the news containing related keywords was relevant to that event.

\subsection{US protests 2017-2018}
We assessed the efficacy of our Graph Language Model (GLM v2) by replicating a real-world scenario. In this scenario, our method was applied to a set of news articles related to the United States, collected over a one-year period from July 2017 to July 2018. Table \ref{res_summ} presents the performance of GLM and Baseline across various events. The 'Event Date' column specifies the dates on which these events occurred. The ‘GLM detected?’ and ‘Baseline detected?’ columns show whether GLM and Baseline, respectively, issued a TP trigger. The ‘GLM trigger date’ and ‘Baseline trigger date’ columns indicate the date of the first window in which each method issued a TP trigger, if any. The ‘GLM Forecast Lead Event’ column displays the number of days in advance GLM issued a TP trigger before the actual event date for single-date events or before the end date for multiple-date events. Finally, the ‘GLM Forecast Lead Baseline’ column indicates how many days in advance GLM issued a TP trigger before the baseline.

\begin{center}

\begin{table}[H]
\small
\centering 
\begin{tabular}
{|p{1.8cm}|p{1.6cm}|p{1.6cm}|p{1.6cm}|p{1.4cm}|p{1.6cm}|p{1.4cm}|p{1.4cm}|} 
\hline
\textbf{Event} & 
\textbf{Event date} & \textbf{GLM  \newline detected?} & \textbf{Baseline \newline detected?} & \textbf{GLM \textit{ } trigger date} & \textbf{Baseline trigger date } & \textbf{GLM \newline Forecast Lead \newline Baseline} & \textbf{GLM\newline Forecast Lead \newline Event} \\
\hline

\textit{Unite the Right rally} & 11-Aug-2017 to 13-Sep-2017 & \textit{Yes} & \textit{Yes} & 14-Aug-2017 & 20-Aug-2017 & 6 & 30\\
\hline
\textit{Gun Control protest} & 24-Mar-2018 & \textit{Yes} & \textit{Yes} & 18-Feb-2018 & 28-Feb-2018 & 10 & 34\\
\hline
\textit{Gaza border protests} & 30-Mar-2018 to 27-Dec-2019 & \textit{Yes} & \textit{No} & 15-May-2018 & - & - & 591\\
\hline
\textit{Immigration policies Protest} & 30-Jun-2018 & \textit{Yes} & \textit{No} & 19-Jun-2018 & 1-Jul-2018 & 12 & 11\\
\hline
\end{tabular}
\caption{\label{res_summ}Results summary US protests}
\end{table}

\end{center}

Table \ref{conf_mat_us} below displays the rates of False Positives (FP), False Negatives (FN), and True Positives (TP). 

\begin{table}[H]
\centering 
\begin{tabular}{|p{3.8cm}||p{1cm}|p{2cm}|p{2cm}|p{2cm}|} 
\hline
\textbf{Model} & \textbf{FP} & \textbf{TP} & \textbf{FN} \\
\hline

\textit{GLM} & \textit{0} & \textit{10} & \textit{2} \\
\hline
\textit{Baseline} & \textit{0} & \textit{10} & \textit{7} \\
\hline

\hline
\end{tabular}
\caption{\label{conf_mat_us}Trigger results for US protests}
\end{table}

In addition, Table \ref{event_mat_us} below displays the number of Misses and Detections:

\begin{table}[H]
\centering 
\begin{tabular}{|p{3cm}||p{1.5cm}|p{2.2cm}|} 
\hline
\textbf{Model} & \textbf{Misses} & \textbf{Detections}  \\
\hline

\textit{GLM} & \textit{0} & \textit{4} \\
\hline
\textit{Baseline} & \textit{2} & \textit{2}  \\
\hline

\hline
\end{tabular}
\caption{\label{event_mat_us} Event results for US protests}
\end{table}

\paragraph{Evaluation per event:}
\begin{enumerate}
   \item \textbf{Unite the Right rally:}
   Our algorithm effectively identified the Unite the Right rally on August 14, 2017, just 3 days after the rally started but 30 days before it concluded, encompassing most of the related protests in that interval. This early detection was based on a dense clique in our graph, including terms like \textit{[‘vigils’, ‘decry white supremacist rally’, ‘protests’]}, which corresponded to the headline: \textit{“Protests, vigils around the US decry white supremacist rally”}. In comparison, the Baseline model detected this event on August 20, 2017, 6 days after GLMv2. Our model’s prediction successfully anticipated a related protest at the University of Virginia on September 13, with further significant trigger points emerging on September 17 and September 19, all pertaining to the same issue.

   
    \item \textbf{Gun Control protest (March for our lives):} Our model successfully forecasted a gun control protest on February 18, 2018, that occurred on March 24, 2018. The prediction was rooted in a news article discussing planned school walkouts and sit-ins after the Florida shooting. This prediction surpassed the baseline, which detected the event on February 28, and also outperformed our earlier prediction from phase 1 (GLM v1 predicted this event on February 22). This improvement may be attributed to the recent modifications we implemented. Figure \ref{clique_gun_control} shows the details of the heaviest clique associated with this event:
    \begin{figure}[H]
    \centering
    \includegraphics[width=0.65\textwidth]{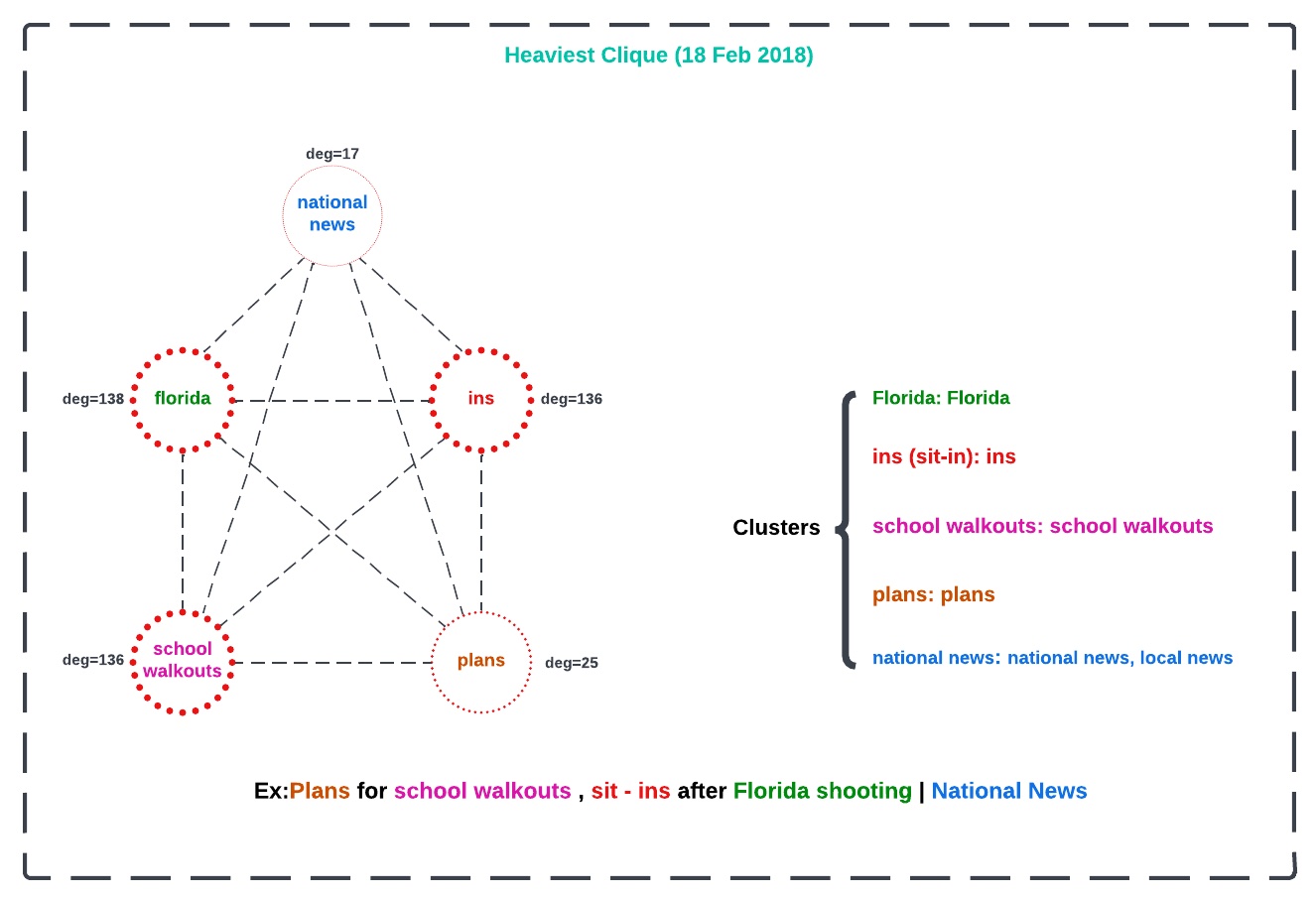}
    \caption{\label{clique_gun_control} Heaviest clique of 18 Feb; 'Degree' indicates the number of news articles containing this specific cluster. Additionally, the size of the dot representing each node is proportional to its degree}
    \end{figure}
    From Figure \ref{clique_gun_control}, it is evident that the heaviest clique comprises nodes such as \textit{['ins', 'school walkouts', 'Florida', 'plans', 'National News']}, with "National News" being the news source.  Notably, the nodes corresponding to sit-ins, school walkouts, and Florida exhibit high degrees within the clique. Upon finding and analyzing the corresponding news, it became apparent that a protest was likely. This exploration aptly demonstrates how our method can indicate the possibility of events.
    \item \textbf{Gaza border protests:} Initiated by President Donald Trump’s decision to relocate the U.S. Embassy to Jerusalem, protests began along the Gaza border on March 30, 2018. Our model recognized these events on May 15, 2018, 10 days prior to their peak escalation. It accurately identified pertinent news headlines, such as ‘The Latest: OIC Condemns US Embassy Move to Jerusalem,’ which were directly linked to the protests. This proficiency enabled us to provide early insights into the major protest that took place on May 25, 2018.
    
    \item \textbf{Immigration policies Protest:} Lastly, we examined a series of protests \textit{against Trump's immigration policies}, wherein our method accurately predicted the most substantial protest in advance. In contrast, the baseline method failed to provide any advanced prediction. A distinguishing aspect of this scenario is our method's ability to leverage data from smaller protests to anticipate a larger, more impactful one.  Figure \ref{clique_immag_prot} depicts the second heaviest clique identified by our algorithm at the point of successful event prediction. Please note that the heaviness of the heaviest clique is not much greater than this one; the difference in heaviness between this and the heaviest clique is less than one. Therefore, we can consider both as important cliques.
   \begin{figure}[H]
    \centering
    \includegraphics[width=0.65\textwidth]{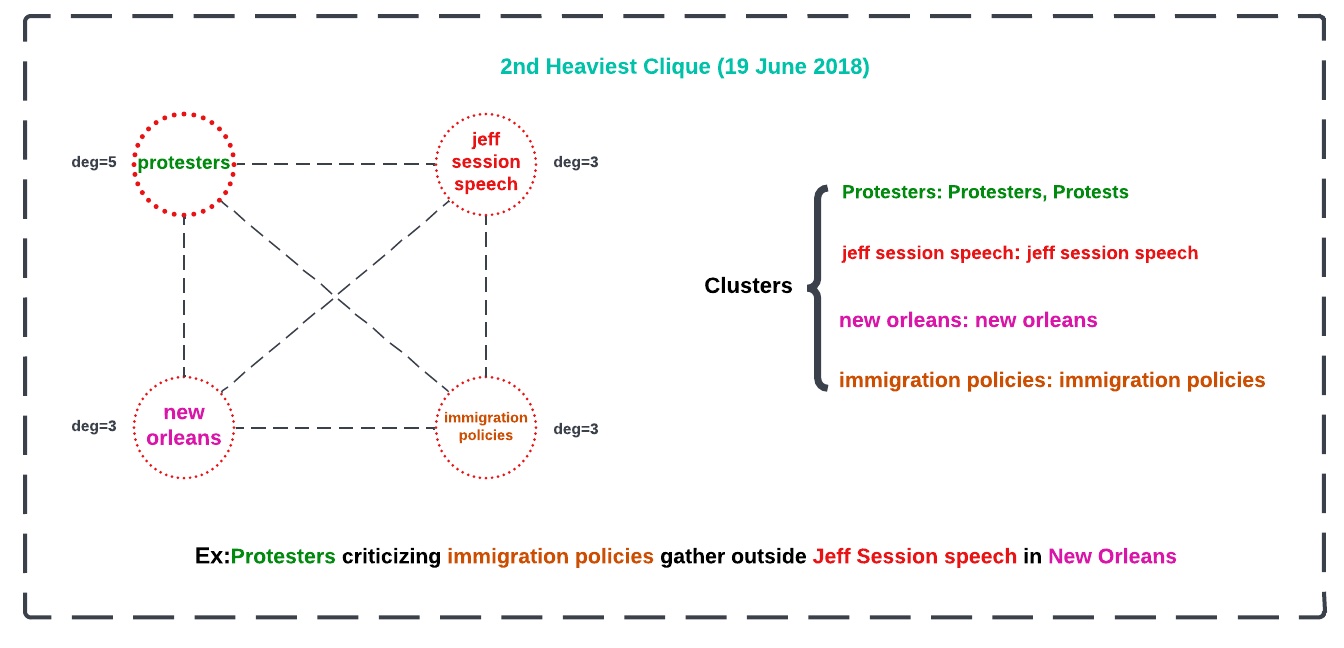}
    \caption{\label{clique_immag_prot} Second Heaviest clique of 19 Jun 2018}
    \end{figure}

\end{enumerate}
\paragraph{Summary:} In summary, our method successfully issued a TP trigger for all 4 events and for 2 of them it occurred before the event starts. For the remaining, the triggers occurred well before the event ends. In contrast, the baseline model managed to detect only two events correctly. Notably, in both instances where the baseline did identify an event, it triggered the detection after our method.

\begin{figure}[H]
\centering
\includegraphics[width=0.9\textwidth]{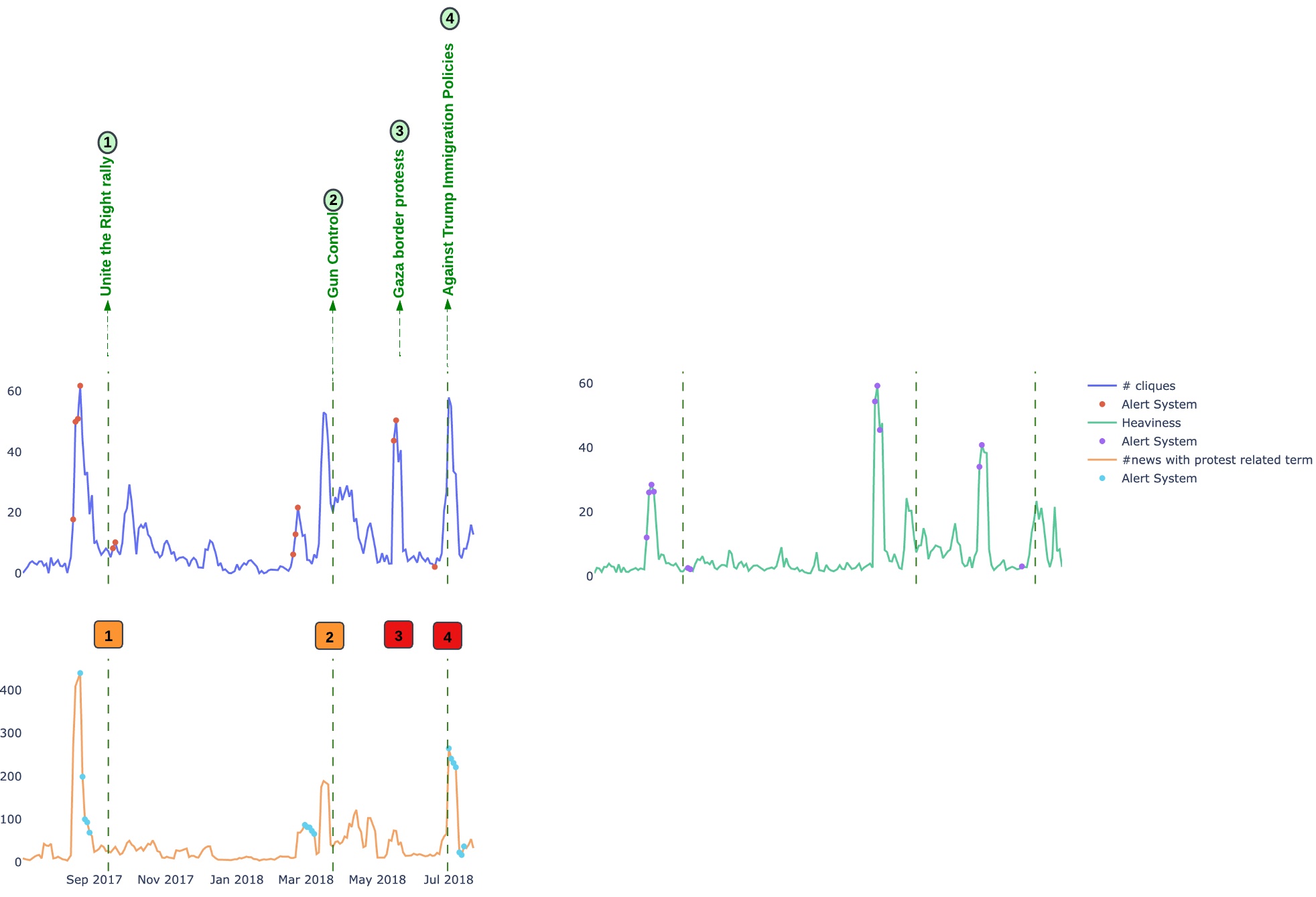}
\caption{\label{US-2017-20182} Representation of all US protests from 2017 to 2018. The numbers display the protests in their chronological order of occurrence, with point markers showing triggers. Color coding is as follows: red indicates protests that the method could not detect; orange signifies protests detected by the method but less effectively than other methods; green represents protests perfectly detected by the method.}
\end{figure}

\subsection{Ukraine War}

\begin{figure}[H]
\centering
\includegraphics[width=0.85\textwidth]{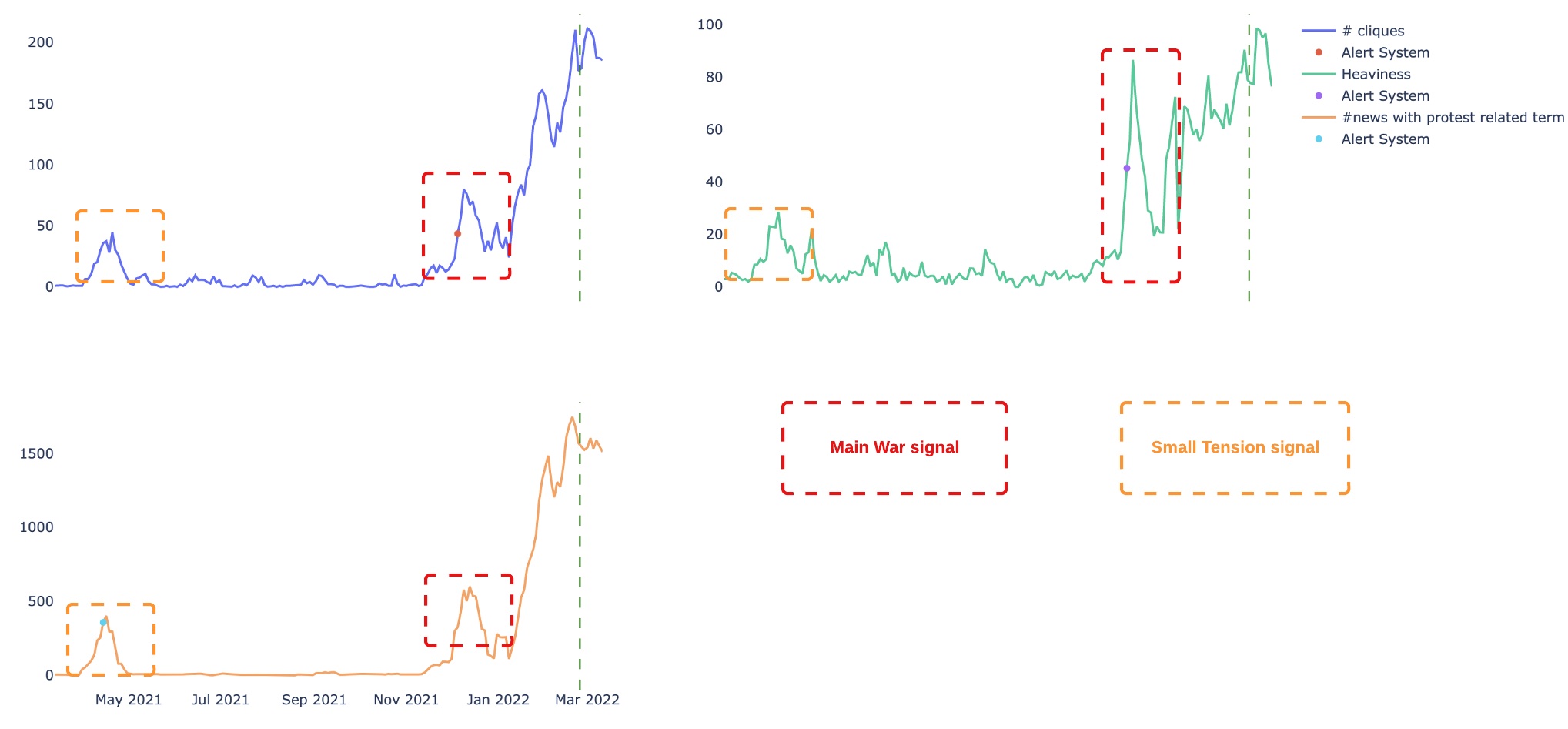}
\caption{\label{war_res_red_orange_alerts} Results of War; Point markers show triggers. The red rectangle highlights high peaks corresponding to the war, visible in both the baseline and GLM, while the orange rectangle indicates peaks associated with minor tensions within the same time interval. The peaks of GLM are notably higher as we approach the war, a trend not mirrored in the baseline. Additionally, we can observe that the baseline only triggers much before the war, unrelated to the war but instead linked to a minor tension. In contrast, the trigger of GLM is a True Positive.}
\end{figure}

Our analysis spanned from March 2021, about a year before the war’s start, to March 2022, shortly after its onset. We applied a higher threshold of $99\%$  to exclude minor tensions within the countries, concentrating on their major conflicts. Our method pinpointed December 5, 2021, as a critical date, with the most significant clique highlighting a news headline about an upcoming Biden-Putin call amid Ukraine tensions. GLMv2 forecasted the event 81 days prior, 8 days sooner than GLMv1. In contrast, the baseline model failed to predict the event. Refer to Figure \ref{war_res_red_orange_alerts} for more details.


\begin{figure}[H]
\centering
\includegraphics[width=0.85\textwidth]{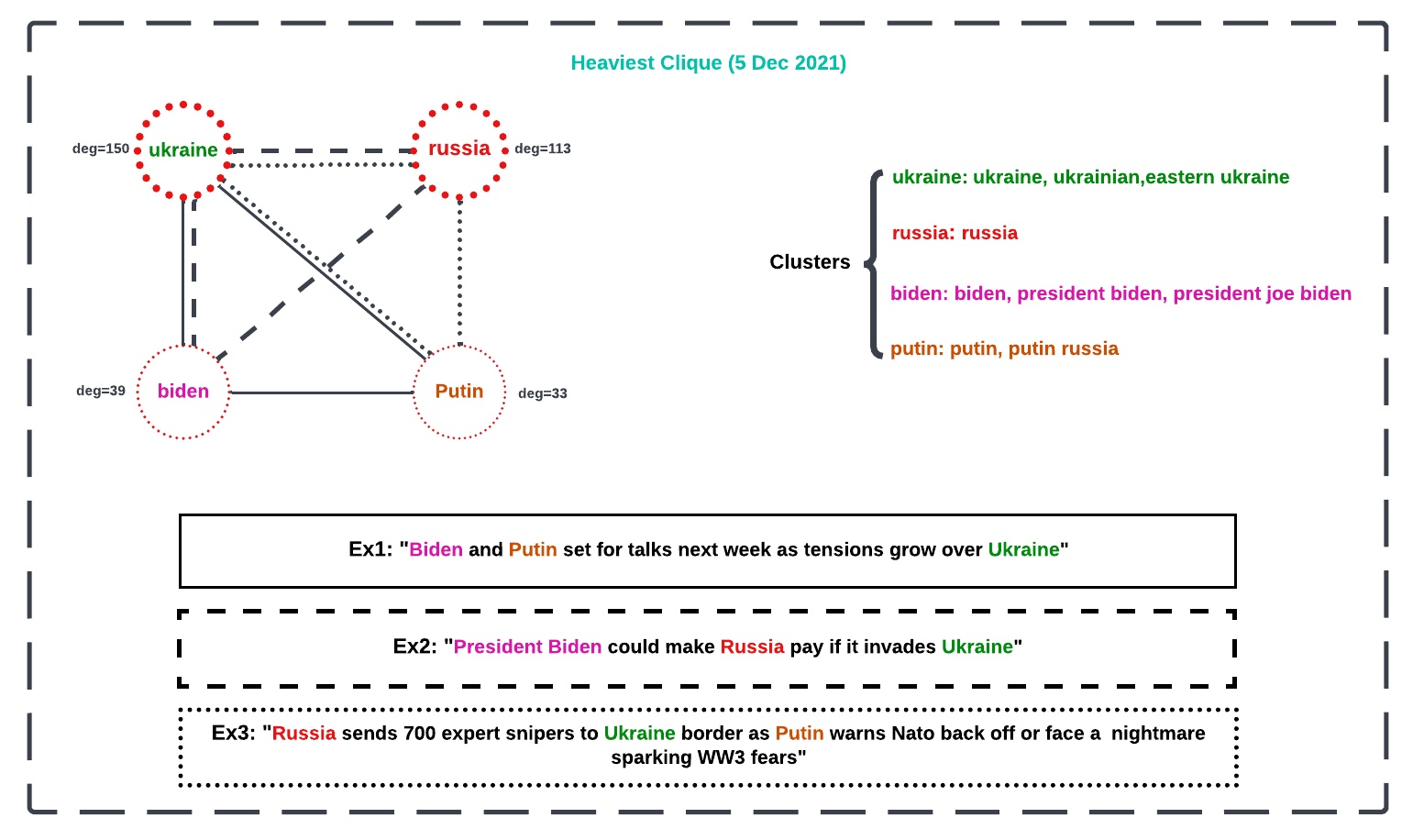}
\caption{\label{clique_war} Heaviest clique of 5 Dec 2021, The shape of the edges indicates headlines that resulted in the formation of those edges.}
\end{figure}
Figure \ref{clique_war} shows which news articles contributed to each edge.  

\paragraph{Summary:} In summary, to encapsulate the findings of this section, the Result Summary (Table \ref{res_summ_war})   and the Trigger results (Table \ref{conf_mat_war}), similar to what we presented for the US protests, are as follows:

\begin{table}[H]
\small
\centering 
\begin{tabular}[H]
{|p{1.6cm}|p{1.4cm}|p{1.7cm}|p{1.7cm}|p{1.5cm}|p{1.5cm}|p{1.5cm}|p{1.5cm}|} 
\hline
\textbf{Event} & 
\textbf{Event date} & \textbf{GLM  \newline detected?} & \textbf{Baseline \newline detected?} & \textbf{GLM  trigger date} & \textbf{Baseline trigger date} & \textbf{GLM \newline Forecast Lead \newline Baseline} & \textbf{GLM \newline Forecast Lead \newline Event} \\
\hline

\textit{Ukraine \newline Russia War} & 24-Feb-2022 & \textit{Yes} & \textit{No} & 5-Dec-2021 & - & - & 81\\
\hline

\end{tabular}
\caption{\label{res_summ_war}Results summary Ukraine War}
\end{table}

\begin{table}[H]
\centering 
\begin{tabular}{|p{3.8cm}||p{1cm}|p{1.5cm}|p{2cm}|p{2cm}|} 
\hline
\textbf{Model} & \textbf{FP} & \textbf{TP} & \textbf{FN} \\
\hline

\textit{GLM} & \textit{0} & \textit{1} & \textit{0} \\
\hline
\textit{Baseline} & \textit{1} & \textit{0} & \textit{0} \\
\hline

\hline
\end{tabular}
\caption{\label{conf_mat_war}Trigger results for Ukraine}
\end{table}

In addition, Table \ref{event_mat_war} below displays the number of Misses and Detections:

\begin{table}[H]
\centering 
\begin{tabular}{|p{3.8cm}||p{1.3cm}|p{2.2cm}|} 
\hline
\textbf{Model} & \textbf{Misses} & \textbf{Detections}  \\
\hline

\textit{GLM} & \textit{0} & \textit{1} \\
\hline
\textit{Baseline} & \textit{1} & \textit{0}  \\
\hline

\hline
\end{tabular}
\caption{\label{event_mat_war} Event results for Ukraine}
\end{table}

\subsection{French Riots}
In recent years, France's diverse range of protests, from opposition to President Macron's pension reforms and cost of living strikes to reactions against police brutality like the Paris shooting and Nahel Merzouk riots, has made it an ideal environment to test our GLM V2 model. We focused on the tumultuous 18-month period from March 2022 to August 2023, characterized by these varied protests alongside major events such as the World Cup and COVID-19 lockdowns. This timeframe allowed us to assess the model's ability to distinguish protest signals in a complex scenario, offering a real-world test of its effectiveness.


Table \ref{res_summ_french} summarizes the detection capabilities of the GLM and Baseline models for each protest, indicating the dates of detection and the lead time of our approach in detecting events earlier.


\begin{table}[H]
\small
\centering 
\begin{tabular}
{|p{2cm}|p{1.6cm}|p{1.6cm}|p{1.6cm}|p{1.5cm}|p{1.5cm}|p{1.4cm}|p{1.5cm}|} 
\hline
\textbf{Event} & 
\textbf{Event date} & \textbf{GLM  \newline detected?} & \textbf{Baseline \newline detected?} & \textbf{GLM \textit{ } trigger date} & \textbf{Baseline trigger date} & \textbf{GLM \newline Forecast Lead \newline Baseline} & \textbf{GLM \newline Forecast Lead \newline Event} \\
\hline

\textit{French Presidential Election Protest} & 16-Apr-2022 & \textit{Yes} & \textit{No} & 11-Apr-2022 & 19-Apr-2022 & 8 & 5\\
\hline

\textit{Cost of living strikes} & 16-Oct-2022 to 10-Nov-2022 & \textit{Yes} & \textit{Yes} & 16-Oct-2022 & 16-Oct-2022 & 0 & 25\\

\hline

\textit{French pension reform unrest(onset)} & 19-Jan-2023 & \textit{Yes} & \textit{No} & 14-Jan-2023 & - & - & 5\\

\hline

\textit{French pension reform unrest(biggest)} & 23-Mar-2023 & \textit{Yes} & \textit{No} & 27-Feb-2023 & - & - & 24\\

\hline

\textit{French pension reform unrest(last)} &  06-Jun-2023  & \textit{Yes} & \textit{No} & 21-Apr-2023 & - & - & 46\\

\hline

\textit{Nahel Merzouk Protests} & 27-Jun-2023 to 15-July-2023  & \textit{Yes} & \textit{Yes} & 30-Jun-2023 & 6-Jul-2023 & 6 & 15\\
\hline
\end{tabular}
\caption{\label{res_summ_french}Results summary France Riots}
\end{table}

Furthermore, Table \ref{conf_mat_frnch} below displays the Trigger results.

\begin{table}[H]
\centering 
\begin{tabular}{|p{3.8cm}||p{1cm}|p{2cm}|p{2cm}|p{2cm}|p{1.5cm}|} 
\hline
\textbf{Model} & \textbf{FP} & \textbf{TP} & \textbf{FN} \\
\hline

\textit{GLM} & \textit{4} & \textit{13} & \textit{1}\\
\hline
\textit{Baseline} & \textit{11} & \textit{4} & \textit{1}\\ 
\hline

\hline
\end{tabular}
\caption{\label{conf_mat_frnch}Trigger results for French Riots}
\end{table}

In addition, Table \ref{event_mat_french} below displays the number of Misses and Detections:

\begin{table}[H]
\centering 
\begin{tabular}{|p{3.8cm}||p{1.3cm}|p{2cm}|} 
\hline
\textbf{Model} & \textbf{Misses} & \textbf{Detections}  \\
\hline

\textit{GLM} & \textit{0} & \textit{6} \\
\hline
\textit{Baseline} & \textit{4} & \textit{2}  \\
\hline

\hline
\end{tabular}
\caption{\label{event_mat_french} Event results for French Riot}
\end{table}


\paragraph{Evaluation per event:}

\begin{enumerate}

\item \textbf{French Presidential Election Protest:}
On April 16, protests occurred in several French cities related to the final stage of the national election. Our method predicted an anomalous point on April 11, 5 days before the event, with the heaviest clique containing \textit{['france', 'pen', 'macron', 'voting']} corresponding to news headline \textit{"A solid showing by French President Macron in the first - round voting bodes well for defeating far-right candidate Marine Le Pen"}. This clique reflected the election competition between Macron and Le Pen. Since the baseline only considers news with explicit protest terms, it did not predict this event in advance. Instead, the baseline detected it on April 19, after protest-related words emerged.

\item \textbf{Cost of living strikes:} In October 2022, strikes related to cost of living erupted in France, with tens of thousands marching in Paris on October 16, 26 days before the end of the event. Both GLM and the baseline correctly detected the onset on October 16. Interestingly, GLM's heaviest clique that day was not relevant. However, the number of related cliques spiked, covering the protest from different angles. By October 18, not only had the number of relevant cliques increased, but the heaviest clique \textit{["france", "thousands", "inflation"]} corresponding to news headline \textit{"Thousands of French people including a Nobel laureate protest over inflation"}also reflected the event.
This example demonstrates how GLM leverages both the number of cliques and heaviness. 
This highlights how GLM captures signals even before a story dominates headlines, via increases in related, lighter cliques.

\item \textbf{French pension reform unrest (onset):} From January to June 2023, widespread protests erupted in France against President Macron's pension reform policies, predominantly led by the General Confederation of Labour (CGT) union. To analyze the performance of our models, we examined three key moments. These were selected as distinct instances for analysis due to the significant scale of each protest, allowing us to consider them separately. The initial wave of unrest emerged on January 19th. Our GLM model detected relevant anomalies on January 14 and 16, suggesting early awareness of the developing situation. The January 14 clique containing \textit{['pension reform', 'strike threat', 'french government']} directly matched the news headline \textit{"French government plays down strike threat over pension reform."}
Thus, GLM detected this protest 5 days early, while the baseline missed it entirely.


\item \textbf{French pension reform unrest (biggest event):} After CGT announced the March 23rd strikes, both models spiked around that date. However, GLM detected earlier signals on February 27th and March 21st, 25 days before the event starts. The heaviest clique on February 27th is \textit{['france', 'pension reform']} corresponding to headline news \textit{"What comes next in France fight over pension reform"}. Though the baseline spiked from March 7 to April 5, it failed to issue a trigger, demonstrating weaker feature extraction.

\item \textbf{French pension reform unrest (last event):} The final significant event in the series of pension reforms occurred on June 6, which GLM successfully anticipated on April 21. On the other end, the baseline did not issue any trigger. The heaviest clique for our GLM detection included \textit{["protests", "french court", "macron plan", "retirement age"]}, aligning with the headline \textit{"French court approves Macron plan to raise retirement age despite protests"}. In contrast, the Baseline model failed to predict this event in advance.

\item \textbf{Nahel Merzouk Protest:} On June 27, 2023, the killing of 17-year-old Nahel Merzouk by police incited protests. The GLM detected this series of events on June 30th, 3 days after its start but 15 days before the end, and 6 days before the baseline, which identified it on July 6th. Analysis of high-degree cliques revealed pertinent nodes about police brutality and teen protests.
In this quickly evolving situation, GLM still surpassed the baseline in terms of detection speed.
Figure \ref{french_clique} presents the details of the heaviest clique.
    \begin{figure}[H]
    \centering
    \includegraphics[width=0.75\textwidth]{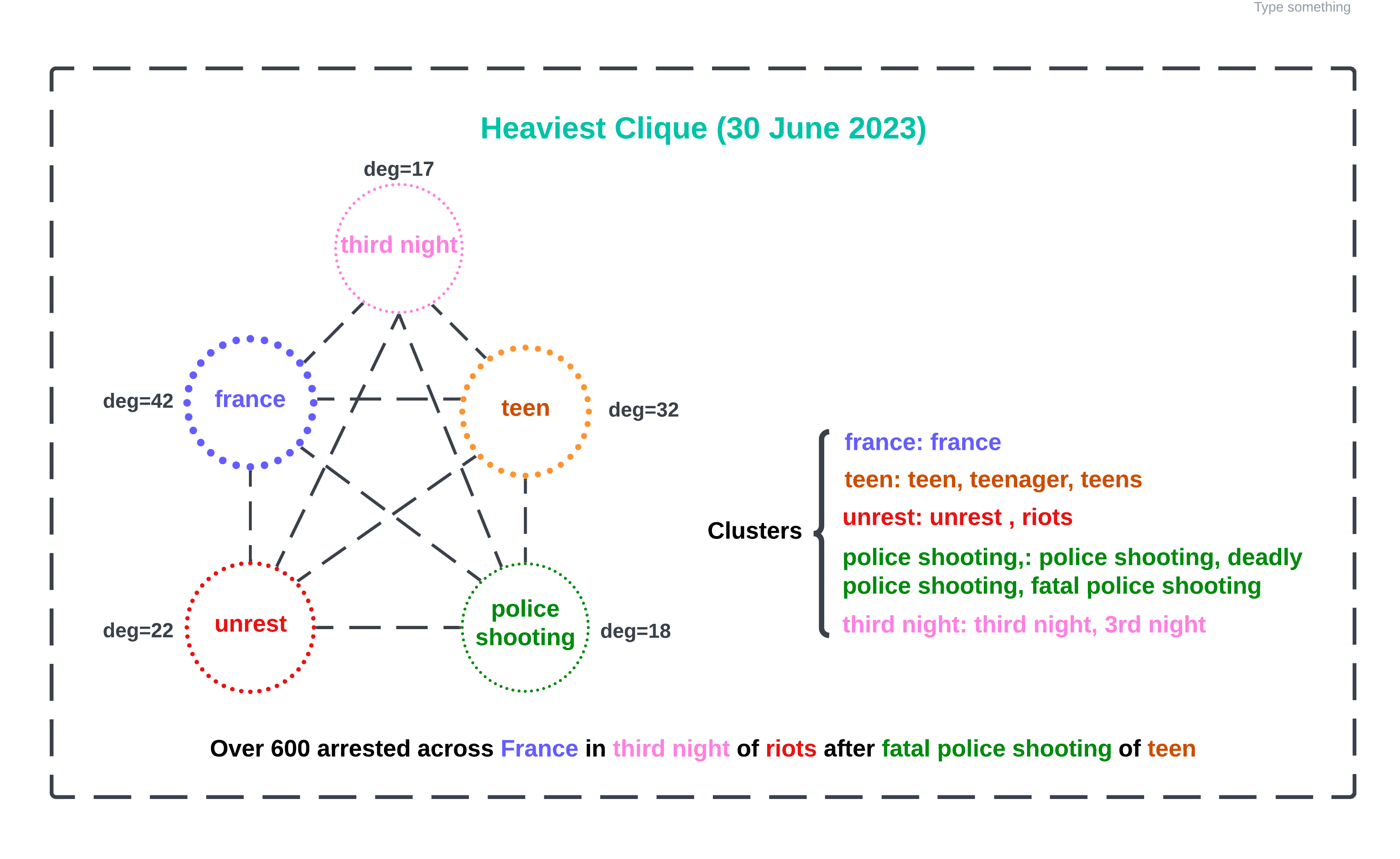}
    \caption{\label{french_clique} The heaviest clique for \textit{Nahel Merzouk Protest} event on 30-Jun-2023 contains nodes such as \textit{['france', 'teen', 'unrest', 'police shooting', 'third night']} ordered by degree. This aligns with headlines like \textit{"Over 600 arrested across France in the third night of riots after the fatal police shooting of the teen."}}
    \end{figure}

\end{enumerate}

\paragraph{Summary:}

In summary, our method successfully issued TP  triggers for all 6 events. For 4 of these events, the triggers occurred before the start of the events, while for the remaining 2, they were well in advance of the event's conclusion. In contrast, the baseline model only managed to detect 2 events correctly. Notably, in one instance where the baseline did detect an event, it triggered the detection after our method.

\begin{figure}[H]
\centering
\includegraphics[width=0.9\textwidth]{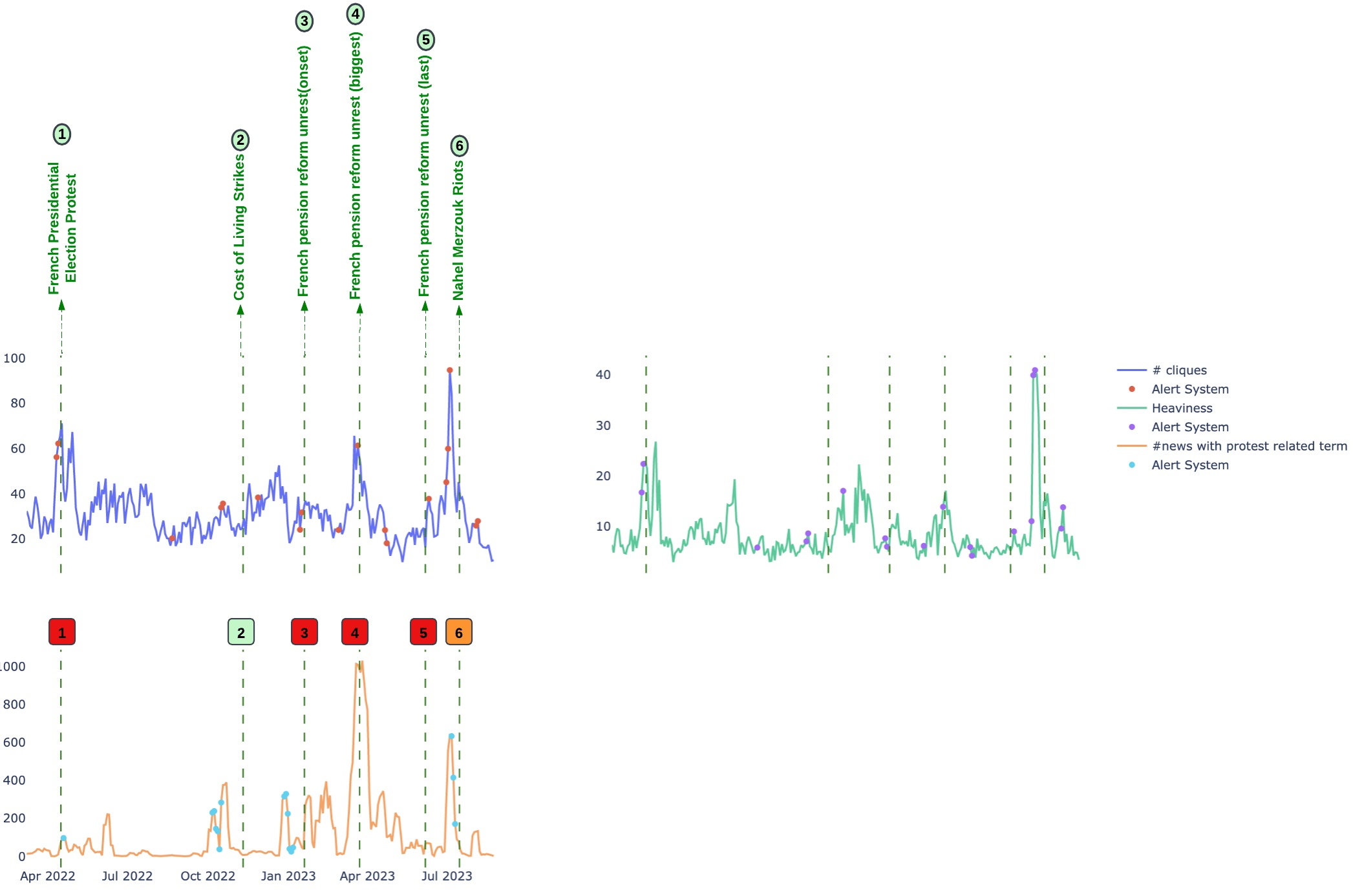}
\caption{\label{US-2017-2018} 
Chronological representation of all French Riots from 2022 to 2023, with point markers showing triggers. Color-coded indicators: Red for undetected protests, Orange for protests detected but later than GLM v2, and Green for protests accurately detected in advance by the method.}
\end{figure}


\section{Conclusion}

GLM v2 consistently surpassed the baseline in the early detection of political events, often predicting them several days ahead of both the baseline model and the actual event date. However, in some instances, while GLM v2 managed to capture the beginning of a series of protests and most of their duration, it was not always able to provide an early sign before they commenced. This is primarily due to two reasons: In cases like the Nahel Merzouk Protest, there were no pre-existing news traces, as the protests were a reaction to an immediate injustice, and journalism requires time to produce content. A potential solution would be to integrate social media as an unsourced and rapid means of disseminating information. Secondly, due to scalability issues, we had to work with a limited amount of data, which reduced the volume of news our algorithm could use to detect anomalies.

The major strengths of our approach over the baseline can be summarized into three primary areas.
\paragraph{Semantic understanding:}
Unlike the baseline's dependence on exact term matches, GLM v2 leverages advanced semantic analysis to understand nuanced language and contextual subtleties. This approach enables it to discern subtle distinctions and relationships between terms, fostering a more profound understanding of complex social phenomena.

\paragraph{Unveiling Contextual Relationships, Beyond Semantics:} GLM v2 excels beyond the baseline model by identifying complex contextual relationships, surpassing basic semantic analysis. While the baseline relies on the frequency of protest-related terms, often leading to delayed recognition of events, GLM v2 adeptly detects nuanced patterns and connections. It scrutinizes the evolving interactions of weighted terms within its graph structure, capturing subtle contextual shifts. This sophisticated approach allows GLM v2 to foresee and distinguish between events at an early stage before they are obvious through term volume alone.


\paragraph{Interpretability:}
The GLM stands out for its ability to provide interpretable insights. Its distinctive structure, characterized by weighted nodes in cliques, skillfully emphasizes crucial themes, prominent entities, and the interplay within communities engaged in emerging protests. Such interpretability is especially beneficial for real-time monitoring, offering a lucid and succinct comprehension of the elements that contribute to early indications in evolving events.

\section{Future Work}
To propel GLM v2 towards its full potential, it is critical to address its current limitations:
\begin{enumerate}
    \item \textbf{Signal Isolation Deficit:} The model currently struggles to identify individual triggering news items, a crucial aspect for precise event analysis.
    \item  \textbf{Lack of Multilingual Support:} GLM v2's effectiveness is confined to a single language, limiting its global applicability and effectiveness in diverse linguistic contexts.

    \item  \textbf{Semantic Limitations:} Despite its already powerful semantic understanding, there is potential for further enhancement in comprehending complex linguistic nuances.

    \item  \textbf{Scalability Issues:} The model faces challenges in handling large data volumes, restricting its usability in data-intensive scenarios.
\end{enumerate}
To move forward, integrating advanced models like LLMs could refine our graph construction methods, and a strategic redesign of our algorithm using approximation techniques along with distributed parallelization may provide a solution to the scalability issues. These steps will be pivotal in unlocking new capabilities and extending the model's reach and efficiency.

\acks{The authors wish to extend their sincere appreciation to Armasuisse for their generous support in funding this research. The opportunity provided by Armasuisse has been invaluable, and the assistance received has greatly contributed to the study's success. We are deeply grateful for their trust and support in our academic endeavors.}

\newpage

\appendix

\section{Implementation}

In this section, we provide a detailed description of the implementation of our graph structures as well as the natural language processing (NLP) components used in our research. We leverage the NetworkX library for graph manipulation and analysis, and various Python packages for NLP tasks. The following subsections provide an overview of each component and its role in our implementation.
\subsection{Downsampling}\label{Downsampling}
To address the time complexity of our algorithms, we implemented downsampling techniques. We introduced a parameter called "freq", which represents a natural number. During each window, we selected only one random news article from every "freq" number of news articles. This downsampling approach allowed us to reduce the computational burden while preserving the overall distribution and relationship of the number of news articles within each window. (For the same reason we perform Filtration after Downsampling)
\subsection{Graph Structures with NetworkX}
For the implementation of our graph structures, we utilize the NetworkX library (\cite{osti_960616}), a powerful Python package that offers a wide range of tools and functionalities for working with graphs. 

 In addition to basic graph functionality, we specifically employed the \textbf{findc cliques} function provided by NetworkX for our signal identification section. By leveraging the \textbf{findc cliques} function, we were able to identify and extract cliques from the graph representation of our data.


\subsection{Natural Language Processing (NLP) Components}

In addition to graph structures, our implementation incorporates various NLP components to preprocess and analyze textual data. The NLP tasks include normalization, text pre-processing, keyword candidate extraction, POS tagging, feature extraction, keyword clustering, anomaly detection, and the creation of interactive dashboards for visualizing time series. The following subsections outline the specific packages and frameworks employed for each task.

\subsubsection{Normalization and Text Pre-processing}
We utilize regular expressions (regex) as a fundamental tool for normalization and text pre-processing. Regex provides a flexible and efficient approach to handle various patterns and structures within the text data, allowing us to clean and standardize the textual content.

\subsubsection{Contextual Filtration}
To filter out news not relevant to politics, we employ a methodology based on contextual similarity assessment. Specifically, we compute the embedding of the word "politics" and compare it to the embedding of news headlines. News headlines with low contextual similarity to the "politics" embedding are filtered out, removing non-political stories. This embedding is computed using the \textit{BAAI/bge-large-en-v1.5} model, which has demonstrated its preeminence by securing the top position in the Massive Text Embedding Benchmark (MTEB) leaderboard. The implementation utilizes the FlagEmbedding library, a versatile tool proficient in mapping textual content to low-dimensional dense vectors. These vectors find application in diverse tasks such as retrieval, classification, clustering, and semantic search.

\subsubsection{Keyword Candidate Extraction and POS Tagging}
To extract keyword candidates and perform Part-of-Speech (POS) tagging, we rely on the Spacy library. Spacy is a popular NLP package in Python that offers efficient and accurate natural language processing capabilities. It provides pre-trained models for various NLP tasks, including POS tagging, which enables us to identify the grammatical components of the text and extract meaningful keywords.

\subsubsection{Sentence Embedding}
For sentence feature extraction, we utilized the SentenceTransformers library with a PyTorch backend. SentenceTransformers is a toolkit that leverages transformer-based models, including BERT and XLM-RoBERTa, to encode sentences into fixed-length feature vectors. Specifically, we employ the state-of-the-art model, \textit{all-mpnet-base-v2}, and leverage it to extract semantic features from news headlines and keyword candidates. These feature representations are essential for the subsequent analysis and clustering tasks.

\subsubsection{Keyword Clustering}
To conduct keyword clustering, we used the \href{https://raw.githubusercontent.com/dmuellner/fastcluster/master/docs/fastcluster.pdf} {fastcluster} algorithm. This algorithm served as a key component in our methodology for clustering important keywords within our dataset. The fastcluster algorithm is a powerful tool for hierarchical clustering, capable of efficiently handling large datasets. By leveraging this algorithm, we were able to group similar keywords together and identify clusters within our data. 

Fastcluster also known as the bottom-up approach or Hierarchical Agglomerative Clustering (HAC) is a type of clustering algorithm that builds a hierarchy of clusters by iteratively merging smaller clusters into larger ones. The algorithm starts with each data point as its own cluster and then proceeds to merge clusters until there is only one cluster left. HAC can be done using various linkage methods, such as single linkage, complete linkage, and average linkage, which determine how the distance between two clusters is calculated, and also the cut-off method is used to determine when to stop the merging process.
To cluster our keywords we used single linkage, cosine similarity as distance matrix (to be more precise 1-cosine similarity), and also considered 0.4 for the cut-off height value.

\begin{figure}[H]
\centering
\includegraphics[width=0.6\textwidth]{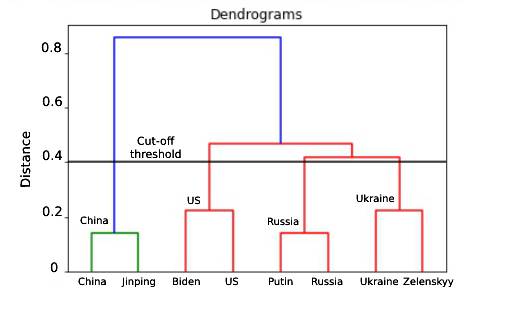}
\caption{\label{fastcluster}The Hierarchical Agglomerative Clustering Dendrogram}
\end{figure}

\paragraph{}
Here is a step-by-step explanation of HAC:

\begin{enumerate}
\item \textbf{Initialize the algorithm:} Start by considering each data point as its own cluster.
\item \textbf{Calculate the distance matrix:} Calculate the distance between each pair of clusters using the single linkage method. Single linkage is the shortest distance between a pair of samples from two clusters A and B, it is given by:

\begin{center}
\begin{math}
d(A, B) = \min  \{ {d(x, y) : x \in A, y \in B } \}
\end{math}
\end{center}

In this formula $d(A, B)$ is the distance between clusters A and B and $d(x,y)$ will be the cosine distance between $x$ and $y$. In other words, we calculate the distances between all pairs of points where one point is from each cluster, and then choose the smallest of these distances. 


\item \textbf{Find the closest clusters:} Identify the two clusters that are closest to each other based on the distance matrix.
\item \textbf{Merge the closest clusters:} Merge the two closest clusters into a new, larger cluster.
\item \textbf{Update the distance matrix:} Recalculate the distances between the new cluster and all the other clusters using the single linkage method, which calculates the minimum distance between elements of each cluster. 
\item \textbf{Repeat steps 3-5 until exceeds cut-off threshold:} Keep merging the closest clusters and updating the distance matrix until a stopping criterion is met. In the cut-off method, the stopping criterion is a pre-defined threshold value for the distance between clusters. Once the distance between the two closest clusters exceeds this threshold, the algorithm stops and returns the clusters obtained at that point as the final result.
\item \textbf{Obtain the final clusters:} The final clusters are the ones obtained at the stopping criterion. They can be represented using a dendrogram (as shown in Figure \ref{fastcluster}), which shows the hierarchy of clusters and the distance (y-axis) at which each merge occurred.
\end{enumerate}

\subsubsection{Anomaly Detection}
For detecting abnormalities within the data, we employ the Isolation Forest (iForest) algorithm available in Scikit-Learn. \cite{scikit-learn}

\subsubsection{Interactive Visualization with Plotly}
To create user-friendly and interactive dashboards for visualizing time series data, we utilize the Plotly library \cite{plotly}. Plotly is a powerful data visualization library in Python that offers a range of charting tools and interactive features. With Plotly, we can create dynamic and intuitive visualizations to explore and present our time series data effectively.

\bibliography{main}
\end{document}